\documentclass[10pt,twocolumn,letterpaper]{article}

\usepackage{cvpr}              

\usepackage{bm}
\usepackage{float}
\usepackage{tcolorbox}
\usepackage{multirow}
\usepackage{colortbl}
\usepackage{tabulary}
\usepackage{etoolbox}
\usepackage{pifont}
\usepackage{makecell}
\usepackage{graphicx}
\usepackage{adjustbox}
\usepackage{amsmath}
\usepackage{amssymb}
\usepackage{booktabs}
\usepackage{rotating}
\usepackage{bbm}
\usepackage{tablefootnote}
\definecolor{LightCyan}{rgb}{0.8,1,1}
\definecolor{LightGreen}{rgb}{0.6,0.9,0.7}
\definecolor{royalblue(traditional)}{rgb}{0.0, 0.14, 0.4}
\definecolor{royalblue(web)}{rgb}{0.25, 0.41, 0.88}
\definecolor{darkgreen}{rgb}{0.0, 0.5, 0.0} 
\definecolor{darkred}{rgb}{0.9, 0.0, 0.0}
\definecolor{darkblue}{rgb}{0.0, 0.0, 0.8}
\usepackage{subcaption}

\newcommand{\cmark}{\textcolor{green!60!black}{\ding{51}}}
\newcommand{\xmark}{\textcolor{red}{\ding{55}}}

\usepackage{tcolorbox}
\usepackage{xcolor}
\usepackage{pgfplots}
\usepackage{algorithm}
\usepackage{algpseudocode}
\usepackage{pgfplotstable}
\usepackage{filecontents}
\usepackage{mdframed}

\usepackage[accsupp]{axessibility}

\definecolor{cvprblue}{rgb}{0.21,0.49,0.74}
\usepackage[pagebackref,breaklinks,colorlinks,allcolors=cvprblue]{hyperref}

\def\paperID{11073} 
\def\confName{CVPR}
\def\confYear{2026}

\title{CLIPoint3D: Language-Grounded Few-Shot Unsupervised 3D Point Cloud Domain Adaptation}

\author{$^{*}\thanks{equal contribution}$Mainak Singha$^{1}$ \and $^{*}$Sarthak Mehrotra$^{2,3}$ \and Paolo Casari$^{1}$ \and Subhasis Chaudhuri$^{3}$ \and \hspace{3cm} Elisa Ricci$^{1,4}$ \and \hspace{1cm} Biplab Banerjee$^{3}$ \hspace{3cm} \and \hspace{-0.5cm}
\small$^{1}$ University of Trento, Italy \and \hspace{-0.75cm} \small$^{2}$ MDSR Labs Adobe, India  \and \hspace{-0.75cm} \small$^{3}$ IIT Bombay, India \and \hspace{-0.75cm} \small$^{4}$ Fondazione Bruno Kessler, Italy
\and
{\tt\small \{mainak.singha, paolo.casari, e.ricci\}@unitn.it}}

\begin{document}
\maketitle
\begin{abstract}
Recent vision-language models (VLMs) such as CLIP demonstrate impressive cross-modal reasoning, extending beyond images to 3D perception. Yet, these models remain fragile under domain shifts, especially when adapting from synthetic to real-world point clouds. Conventional 3D domain adaptation approaches rely on heavy trainable encoders, yielding strong accuracy but at the cost of efficiency. We introduce \textbf{CLIPoint3D}, the first framework for \textit{few-shot unsupervised 3D point cloud domain adaptation} built upon CLIP. Our approach projects 3D samples into multiple depth maps and exploits the frozen CLIP backbone, refined through a \textit{knowledge-driven prompt tuning} scheme that integrates high-level language priors with geometric cues from a lightweight 3D encoder. To adapt task-specific features effectively, we apply \textit{parameter-efficient fine-tuning} to CLIP's encoders and design an \textit{entropy-guided view sampling} strategy for selecting confident projections. Furthermore, an \textit{optimal transport-based alignment loss} and an \textit{uncertainty-aware prototype alignment loss} collaboratively bridge source-target distribution gaps while maintaining class separability. Extensive experiments on \textbf{PointDA-10} and \textbf{GraspNetPC-10} benchmarks show that {CLIPoint3D} achieves consistent 3-16\% accuracy gains over both CLIP-based and conventional encoder-based baselines. Project page: \url{https://sarthakm320.github.io/CLIPoint3D}.

\end{abstract}    
\section{Introduction}
\label{sec:intro}

Point cloud understanding underpins modern 3D vision, driving applications in autonomous driving~\cite{3dautodriving}, terrain mapping~\cite{3dterrain}, augmented reality, and robotics~\cite{3drobotics}. Unlike 2D imagery, point clouds explicitly encode fine-grained geometric cues essential for spatial reasoning. Despite the remarkable progress of deep 3D architectures~\cite{pointnet,dgcnn,spatio3d}, most assume identical training and deployment distributions. In practice, scans acquired from heterogeneous sensors exhibit large variations in point density, sampling patterns, occlusion, and background clutter, leading to severe performance degradation under domain shifts~\cite{pcchallenges,pcgeoreferenced}. This issue is exacerbated when transferring from synthetic benchmarks to real-world environments, making \textit{unsupervised domain adaptation (UDA)}~\cite{uda,mmd, dann, cdan, acda, scda, cosmo} central to achieving scalable 3D perception.

\begin{figure}[t]
    \centering
    \includegraphics[width=1.02\columnwidth]{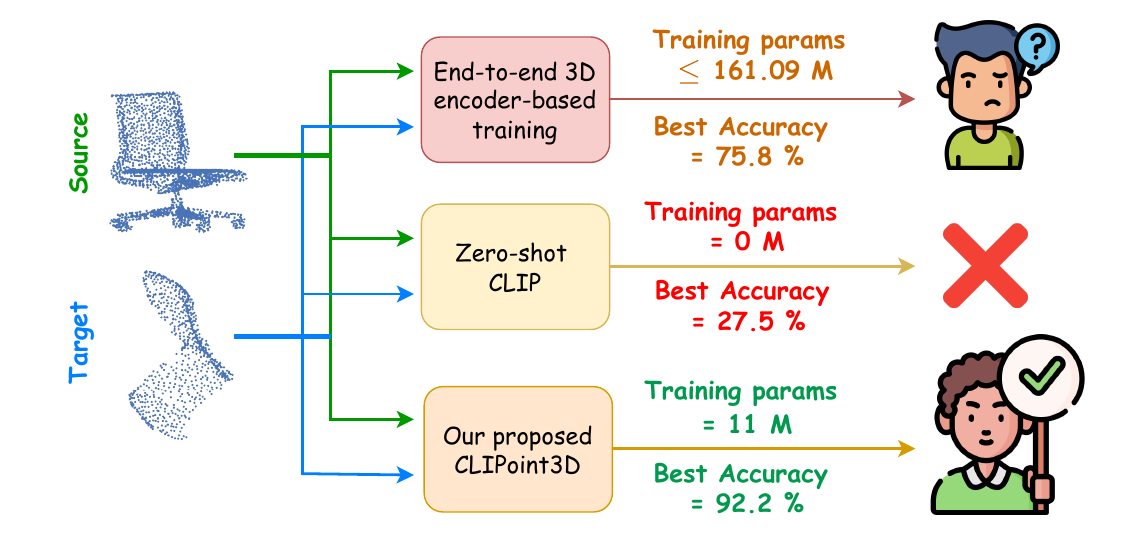}
    \caption{\textbf{Comparison of CLIPoint3D with SOTA methods on GraspNetPC-10.}  
    Encoder-based 3D UDA methods (e.g., PointDAN~\cite{pointdan}, GAST~\cite{gast}, MLSP~\cite{mlsp}) are accurate but computationally expensive, while CLIP-based extensions fail to bridge the synthetic-real gap.  
    CLIPoint3D achieves \textit{+16.4\%} improvement with minimal overhead.}
    \label{fig:teaser}
\end{figure}

Direct fine-tuning on target data can partially mitigate domain gaps but requires dense 3D annotations and high compute, impractical for dynamic or safety-critical applications~\cite{dabudget,dgsurvey}. Since 3D labeling is costly and error-prone~\cite{resource3d}, UDA methods aim to transfer knowledge from labeled sources to unlabeled targets. The key difficulty lies in jointly enforcing \textit{statistical alignment} (distributional consistency) and \textit{semantic alignment} (class-level coherence); neglecting either leads to geometrically aligned yet semantically inconsistent features.

However, existing unsupervised point cloud domain adaptation (UPDA) techniques generally fall into three paradigms.  
(\textit{i})~\emph{Adversarial alignment}~\cite{dann,pointdan} uses domain discriminators to match latent features but often suffers from model collapse and over-alignment. (\textit{ii})~\emph{Self-supervised learning}~\cite{rs,defrec,3denet} employs pretext tasks such as rotation or deformation prediction, but lacks semantic awareness. (\textit{iii})~\emph{Pseudo-labeling and self-paced learning}~\cite{gast,gai,mlsp} iteratively refine noisy labels, yet degrade under large shifts. Although effective in controlled setups, these models are geometry-centric, computationally heavy, and rarely leverage semantic priors or uncertainty estimation, limiting their robustness to unseen modalities.

Recently, vision-language models (VLMs) such as CLIP~\cite{clip} have demonstrated impressive zero-shot transfer by coupling visual and textual modalities through large-scale contrastive pretraining. Extending CLIP to 3D~\cite{pointclip,pointclipv2,clip2point,diffclip,clipgoes3d} typically involves projecting point clouds into multi-view depth maps and processing them via CLIP’s image encoder. While effective for single-domain tasks, such projections expose two fundamental limitations: (\textit{i})~\textbf{Modality gap:} CLIP’s encoder, trained on RGB images, poorly captures the sparse, textureless, and geometry-dominant nature of 3D depth maps;  
(\textit{ii})~\textbf{Domain gap:} Existing CLIP-3D models~\cite{c3pr,filp3d,cmgr} lack mechanisms for cross-domain adaptation, yielding poor generalization beyond their source domain. The following issues raise a crucial research question: \textit{How can we harness CLIP’s semantic priors to enable unsupervised 3D domain adaptation while bridging both the 2D-3D modality and source-target domain gaps in a compute-efficient way?}

We hypothesize that CLIP’s language-grounded latent space can be effectively adapted for 3D UDA if jointly guided by geometric cues and uncertainty-aware optimization. This motivates a framework that (i) injects geometric awareness into CLIP’s latent space, (ii) aligns distributions across domains without labels, and (iii) achieves parameter-efficient adaptation for few-shot supervision.

\textbf{Our Approach.}  
We introduce \textbf{CLIPoint3D}, a unified framework for \textit{few-shot unsupervised 3D domain adaptation} built on top of CLIP. It projects each point cloud into multiple depth maps~\cite{pointclip} and reuses CLIP’s frozen visual backbone, leveraging 2D pretraining for efficient 3D transfer. A \textit{knowledge-driven prompt tuning} module fuses high-level semantic priors from large language models (LLMs)~\cite{gpt,phi4,llama3,qwen2,gpt5} with low-level geometric features from a lightweight 3D encoder, grounding CLIP’s embeddings in 3D structure.  
To further reduce compute, we employ \textit{parameter-efficient fine-tuning (PEFT)}~\cite{peft} to adapt a small subset of CLIP parameters while preserving its zero-shot capability. An \textit{entropy-guided view sampling} strategy~\cite{tpt} filters ambiguous or redundant views to stabilize multi-view aggregation. Finally, we introduce two \emph{novel} alignment objectives: (i) an \textbf{uncertainty-aware prototype alignment loss} that performs class-level coupling using entropy-weighted prototypes, and (ii) an \textbf{entropy-regularized OT alignment} that enforces smooth, noise-tolerant global matching.  
Together, these confidence-aware objectives yield robust semantic and distributional alignment under large 3D domain shifts.

In summary, our key contributions are:

\begin{enumerate}[leftmargin=0.6cm]
\item The first CLIP-based framework for few-shot unsupervised 3D point cloud domain adaptation, achieving strong cross-domain generalization with minimal training cost.

\item A knowledge-driven prompt tuning scheme that unites LLM-derived semantic priors and 3D geometry for multimodal grounding. 

\item Dual uncertainty-aware objectives, OT-based statistical alignment and prototype-level semantic regularization, that tighten the adaptation generalization bound (Sec.~\ref{sec:gen_bound}).

\item An entropy-guided view selection mechanism that enhances stability, interpretability, and efficiency under multi-view uncertainty.
\end{enumerate}

Extensive experiments and ablations across standard benchmarks demonstrate that CLIPoint3D achieves superior accuracy-efficiency trade-offs compared to both 3D encoder-based and CLIP-based baselines (Figure ~\ref{fig:teaser}).

\begin{figure*}[htbp]
    \centering
    \includegraphics[width=1.02\textwidth]{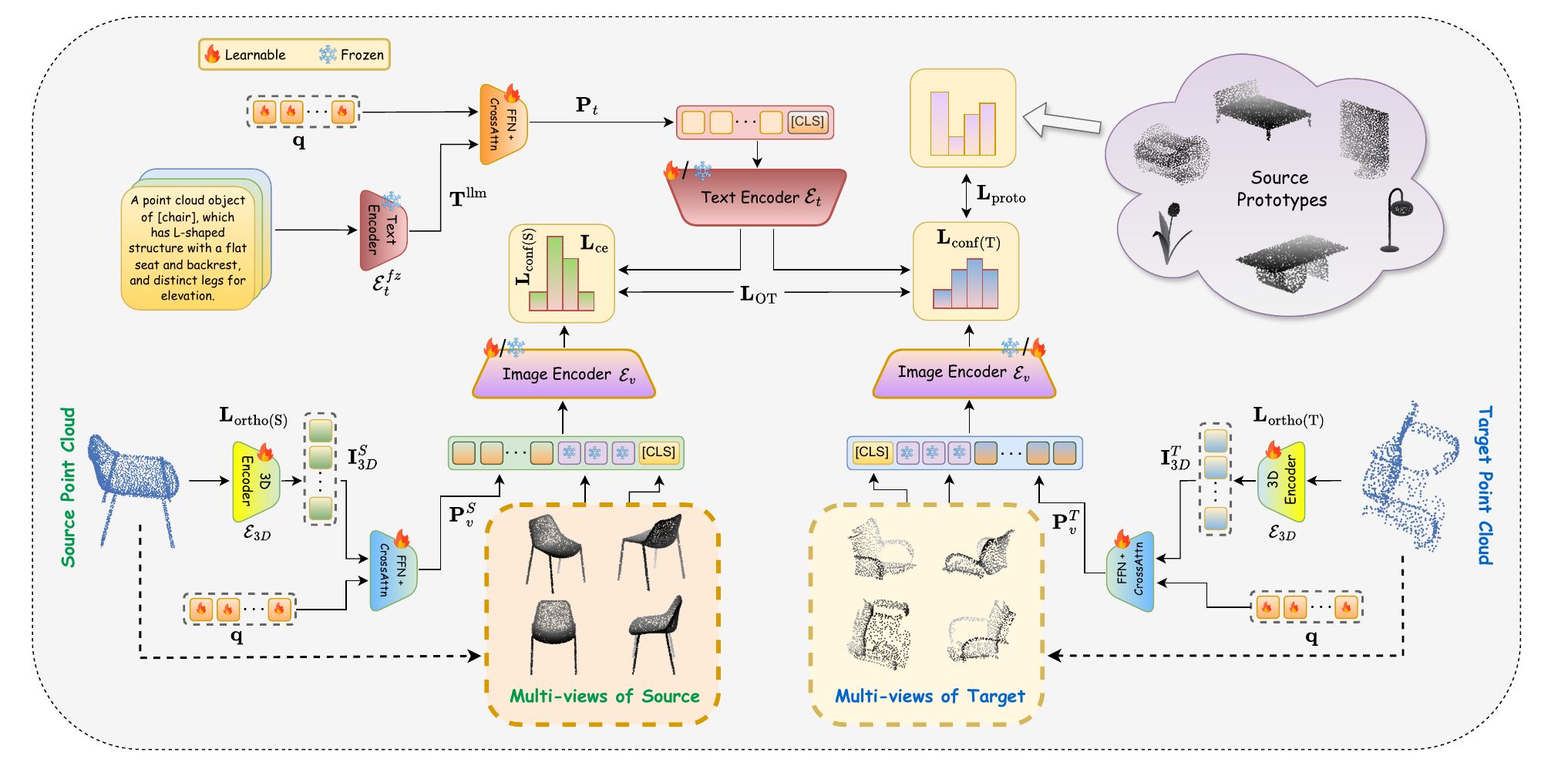}
    \caption{\textbf{Overview of CLIPoint3D}, the first CLIP-based unsupervised 3D point cloud domain adaptation framework, comprises four key modules: (1)~\textit{Knowledge-driven prompt tuning} generates LLM-guided textual and 3D-aware visual prompts;  
    (2)~\textit{Parameter-efficient fine-tuning (PEFT)} jointly optimizes these prompts and the encoder while (3) \textit{entropy-based view selection} filters unreliable projections;  
    (4)~Dual objectives, uncertainty-aware prototype loss $\mathbf{L}_{\mathrm{proto}}$ and optimal transport loss $\mathbf{L}_{\mathrm{OT}}$, achieve joint semantic and statistical alignment.  
    Additional regularizers include $\mathbf{L}_{\mathrm{conf}} = \mathbf{L}_{\mathrm{conf(S)}} + \mathbf{L}_{\mathrm{conf(T)}}$, and $\mathbf{L}_{\mathrm{ortho}} = \mathbf{L}_{\mathrm{ortho(S)}} + \mathbf{L}_{\mathrm{ortho(T)}}$, to ensure stable learning across source and target domains.}
    \label{fig:main_fig}
    \vspace{-0.35cm}
\end{figure*}

\section{Related Works}
\label{sec:related_works}

\noindent\textbf{Unsupervised 3D Point Cloud Domain Adaptation.}
UPDA seeks to transfer knowledge from labeled source to unlabeled target domains. Existing approaches mainly follow three paradigms.  
(\textit{i})~\textit{Domain adversarial training}~\cite{dann,pointdan} employs discriminators to enforce feature invariance while the feature extractor learns to confuse them. Although conceptually sound, such methods often suffer from unstable convergence, mode collapse, and over-alignment that compromises geometric fidelity critical for fine-grained 3D recognition.  
(\textit{ii})~\textit{Self-supervised learning (SSL)}~\cite{rs,defrec,3denet} leverages auxiliary pretext tasks such as rotation prediction~\cite{gast} or deformation reconstruction~\cite{defrec} to capture domain-invariant cues. While interpretable, SSL mainly learns low-level geometric invariances with limited semantic alignment.  
(\textit{iii})~\textit{Pseudo-label and self-paced learning}~\cite{gast,gai,mlsp} iteratively refines pseudo labels for confident target samples, but noisy labels under large shifts amplify confirmation bias.  
Overall, most UPDA methods are geometry-centric, computationally heavy, and lack semantic grounding or uncertainty modeling, making them less effective for lightweight few-shot adaptation.

\noindent\textbf{CLIP for 3D Understanding.}
Large-scale vision-language models such as CLIP~\cite{clip} learn rich multimodal embeddings by aligning image and text representations, inspiring a range of 3D extensions. PointCLIP~\cite{pointclip}, PointCLIP v2~\cite{pointclipv2}, DiffCLIP~\cite{diffclip}, and MVFPoint~\cite{mvfpoint} project 3D point clouds into multi-view depth maps and process them via CLIP’s image encoder, achieving strong zero-/few-shot classification. CG3D~\cite{clipgoes3d} aligns point cloud-image-text triplets through visual prompt tuning to reduce the RGB-depth gap, while $\text{CLIP}^2$~\cite{clip2} leverages proxy alignment from 2D-3D correspondences for transferable 3D features. Few-shot class-incremental frameworks~\cite{c3pr,filp3d,cmgr} further exploit CLIP’s semantics to reprogram depth projections for within- and cross-domain generalization.  
However, these works primarily focus on recognition rather than adaptation. They often freeze CLIP encoders, lacking explicit mechanisms for cross-domain alignment or handling uncertain multi-view projections, limiting robustness under shifts.

\noindent\textbf{CLIP for 2D UDA.} Recent studies have successfully adapted CLIP for 2D unsupervised domain adaptation. DAPL~\cite{dapl} introduces pseudo-labeling for target samples, whereas AD-CLIP~\cite{adclip} aligns source and target domains in the textual prompt space while preserving style semantics. PADCLIP~\cite{padclip} proposes an adaptive debiasing pseudo-labeling strategy based on forgetting measures. UniMoS~\cite{unimos} employs modality-ensemble training to balance modality-agnostic and modality-specific features. DAMP~\cite{damp} jointly aligns visual and textual embeddings to enhance domain invariance, while PDB~\cite{pdb} decomposes adaptation into multiple sub-tasks using auxiliary data construction and cascaded semantic filters, demonstrating CLIP's versatility for 2D adaptation and its ability to leverage cross-modal representations to improve generalization to target domains. However, these methods are inherently designed for RGB images and do not account for the unique geometric challenges of 3D data explicitly.

\section{Proposed Methodology}
\label{sec:methodology}

The UPDA setup involves a labeled source domain and an unlabeled target domain.  
The source set $\mathcal{D}_S = \{(\mathbf{PC}_i^S, y_i^S)\}_{i=1}^{N_S}$ contains $N_S$ labeled point clouds, while the target set $\mathcal{D}_T = \{\mathbf{PC}_j^T\}_{j=1}^{N_T}$ includes $N_T$ unlabeled ones.  
Each 3D point cloud $\mathbf{PC}$ is projected into $M$ depth views, yielding  
$\mathcal{D}_S = \{(x_{i,m}^S, y_i^S)\}$ and $\mathcal{D}_T = \{x_{j,m}^T\}$,  
where $x_{i,m}^S, x_{j,m}^T$ denote the $m$-th 2D projections and $m \!\in\! \{1,\ldots,M\}$.  
Samples follow distinct distributions $\mathcal{P}_S$ and $\mathcal{P}_T$ ($\mathcal{P}_S \!\neq\! \mathcal{P}_T$) but share a common label space $\mathcal{Y}$.  
The goal is to learn a classifier  
$f\!:\!\mathcal{X}_S \!\rightarrow\! \mathcal{Y}$  
that generalizes to $\mathcal{X}_T$ by jointly leveraging $\mathcal{D}_S$ and $\mathcal{D}_T$ in a transductive manner.

UPDA remains challenging due to:  
(i)~large geometric and density variations across sensors and domains;  
(ii)~information loss and redundancy from 3D-to-2D projection, complicating cross-view consistency;  
(iii)~absence of target labels, which blurs the boundary between semantic drift and domain shift; and  
(iv)~the limited transferability of 2D VLMs like CLIP, pretrained on textured RGB data, to sparse, textureless 3D projections.

\subsection{Our CLIPoint3D Framework}
\label{sec:method}

{CLIPoint3D} builds upon a frozen CLIP backbone composed of a vision encoder $\mathcal{E}_v$ and a text encoder $\mathcal{E}_t$. Following \cite{pointclip}, we employ online perspective projection \cite{projection} without any post-rendering operations \cite{postrendering}, directly projecting each 3D point onto a set of predefined image planes to produce scatter-based depth maps. For each projected view $x_{i,m}$ of a 3D point cloud $\mathbf{PC}_i$, the vision encoder produces an embedding  
$\mathbf{v}_{i,m} = \mathcal{E}_v(x_{i,m}) \in \mathbb{R}^{1 \times d}$,  
where $d$ is the feature dimension.  
For $K$ categories, text templates are encoded as  
$\mathbf{T} = \mathcal{E}_t(\mathbf{t}) \in \mathbb{R}^{K \times d}$,  
with $\mathbf{t}$ denoting the set of class templates.  
Given temperature $\tau$, the probability of assigning class $y$ to view $x_{i,m}$ is:
\begin{equation}
p(y|x_{i,m}) =
\frac{\exp\!\left(\cos(\mathbf{v}_{i,m}, \mathbf{T}_y)/\tau\right)}
{\sum_{k=1}^{K}\exp\!\left(\cos(\mathbf{v}_{i,m}, \mathbf{T}_k)/\tau\right)}.
\label{eq:clip_prob}
\end{equation}

To effectively adapt CLIP to 3D understanding under domain shift, {CLIPoint3D} refines its latent space through four complementary modules (in Figure \ref{fig:main_fig}):  
(i)~\textit{Knowledge-driven prompt tuning}, integrating LLM-guided language priors with geometric cues for coherent adaptation of both encoders;  
(ii)~\textit{PEFT}, selectively updating a minimal set of CLIP's parameters to specialize on multi-view 3D structure;  
(iii)~\textit{Entropy-guided view selection}, filtering uncertain projections to enhance feature consistency; and  
(iv)~\textit{Uncertainty-aware prototype alignment}, aligning source-target domains while preserving class separability.

\subsubsection{Knowledge-Driven Prompt Tuning}
\label{sec:kpt}

Conventional prompt-tuning methods~\cite{factualtuning,prefixtuning,paramtuning} adapt pretrained models by inserting lightweight learnable tokens while keeping the backbone frozen. Techniques such as CoOp~\cite{coop}, VPT~\cite{vpt}, and MaPLe~\cite{maple} work well for 2D images because texture strongly correlates with semantics. However, when transferred to sparse 3D projections, these assumptions fail: (i) textual prompts remain purely linguistic without geometric grounding, and (ii) visual prompts depend on texture cues that are absent in point clouds. Consequently, CLIP’s latent space remains biased toward its 2D pretraining distribution and struggles to represent 3D data.

To address this limitation, we propose a \textit{knowledge-driven multimodal prompt-tuning} strategy that links linguistic semantics with 3D structure. We jointly adapt CLIP’s text and vision encoders using two complementary knowledge sources: (a) high-level semantic priors from an LLM~\cite{gpt5}, and (b) low-level geometric descriptors from a lightweight 3D encoder $\mathcal{E}_{3D}$~\cite{pointnet}. A shared query vector $\mathbf{q}$ drives cross-modal attention, ensuring that both textual and visual prompts evolve around a common semantic reference while retaining modality-specific flexibility. This yields geometry-aware semantic reasoning and stable domain transfer.

\textbf{$\boldsymbol{\rightarrow}$ Textual Prompt Generation.}
For each class label $y_k \!\in\! \mathcal{Y}$, the LLM generates a descriptive sentence 
\texttt{``a 3D point cloud object of a [CLS] with [attributes]''}, 
anchoring the prompt explicitly within the 3D modality. The frozen CLIP text encoder $\mathcal{E}_{t}^{fz}$ (where $fz$ denotes frozen) encodes these phrases as:
\begin{equation}
\mathbf{T}^{\text{llm}} = 
\{\mathcal{E}_{t}^{fz}(\texttt{prefix}+\texttt{LLM}(y_k))\}_{k=1}^{K}.
\end{equation}
A text-side multi-head cross-attention (MHCA) module then refines these embeddings using the shared query $\mathbf{q}$:
\begin{equation}
\mathbf{P}_t =
\text{FFN}\!\big(\text{MHCA}(\mathbf{Q}_{\mathbf{q}},
\mathbf{K}_{\mathbf{T}^{\text{llm}}},
\mathbf{V}_{\mathbf{T}^{\text{llm}}})\big),
\end{equation}
where $\mathbf{Q}_{\mathbf{q}} = \mathbf{q}W_{\mathbf{Q}}$,  
$\mathbf{K}_{\mathbf{T}^{\text{llm}}}=\mathbf{T}^{\text{llm}}W_{\mathbf{K}_t}$, and  
$\mathbf{V}_{\mathbf{T}^{\text{llm}}}=\mathbf{T}^{\text{llm}}W_{\mathbf{V}_t}$.  
Here, FFN denotes a feed-forward network.  
The learned textual prompt $\mathbf{P}_t$ conditions the text encoder $\mathcal{E}_t$, producing geometry-aware, domain-stable embeddings  $\mathbf{T} = \mathcal{E}_t(\mathbf{t}, \mathbf{P}_t)$.

\textbf{$\boldsymbol{\rightarrow}$ Visual Prompt Generation.}
Given a 3D point cloud $\mathbf{PC}$, its structural feature representation  $
\mathbf{I}_{3D} = \mathcal{E}_{3D}(\mathbf{PC})
$ 
is injected into a parallel vision-side MHCA block using the same shared query $\mathbf{q}$:
\begin{equation}
\mathbf{P}_v =
T_{\text{proj}}\!\big(
\text{FFN}\!\big(\text{MHCA}(\mathbf{Q}_{\mathbf{q}},
\mathbf{K}_{\mathbf{I}_{3D}},
\mathbf{V}_{\mathbf{I}_{3D}})\big)\big),
\end{equation}
where  
$\mathbf{K}_{\mathbf{I}_{3D}}=\mathbf{I}_{3D}W_{\mathbf{K}_v}$  
and  
$\mathbf{V}_{\mathbf{I}_{3D}}=\mathbf{I}_{3D}W_{\mathbf{V}_v}$.  
The projection layer $T_{\text{proj}}$ maps $\mathbf{P}_v$ to CLIP’s patch-embedding dimension, producing geometry-aware visual prompts $\mathbf{P}_v^{S}$ and $\mathbf{P}_v^{T}$ for source and target domains, respectively. Distinct parameter sets $(W_{\mathbf{K}_t},W_{\mathbf{V}_t})$ and $(W_{\mathbf{K}_v},W_{\mathbf{V}_v})$ ensure modality-specific adaptability, while the shared query vector $\mathbf{q}$ maintains a unified alignment objective.

By fusing LLM-driven semantic hierarchies with invariant 3D structural priors under a shared attention query,  
{CLIPoint3D} transforms visual prompt tuning from shallow token reweighting into structured multimodal knowledge transfer. The visual embedding for a projected view of a specific domain becomes  $\mathbf{v}_{i,m} = \mathcal{E}_v(x_{i,m}, \mathbf{P}_v)$.
This yields three notable benefits:  
(i) enhanced text–vision correspondence under missing appearance cues,  
(ii) stable cross-domain adaptation through geometry-grounded semantics, and  
(iii) parameter-efficient adaptation with minimal computational overhead (see Table~\ref{tab:prompting}).

\subsubsection{Few-Shot PEFT Adaptation}
\label{sec:peft}

While knowledge-driven prompt tuning aligns linguistic and geometric cues, CLIP’s encoders, pretrained on RGB images and short captions, still struggle to model fine-grained 3D structure. Fully fine-tuning $\mathcal{E}_v$ on sparse projections risks overfitting and disrupting the pretrained image-text alignment crucial for generalization. To enable targeted geometric adaptation without harming semantic consistency, {CLIPoint3D} employs a LoRA-based PEFT~\cite{lora}.

$\mathcal{E}_v$ is decomposed into the frozen backbone and lightweight low-rank adapters that capture 3D-specific residual cues such as curvature, surface continuity, and depth transitions. Updating only these adapters stabilizes gradients, reduces parameter cost, and avoids drifting from CLIP’s semantic priors while still modeling domain-dependent geometric variations. We also apply PEFT to $\mathcal{E}_t$, whose pretrained space is biased toward 2D natural-image semantics. Low-rank adapters introduce controlled shifts that align LLM-enhanced prompts with 3D structural attributes without perturbing global CLIP alignment. Table~\ref{tab:peft} shows complementary gains from adapting each branch individually and jointly.

Thus, PEFT serves as the \emph{local adaptation layer} of {CLIPoint3D}: prompt tuning provides global semantic alignment, while PEFT refines encoder features for domain- and task-specific structure, ensuring the model remains both semantically coherent and geometrically discriminative under few-shot supervision.

\subsubsection{Entropy-Guided View Selection}
\label{sec:entropy_view}

In parallel to multimodal prompt tuning and PEFT adaptation, each point cloud $\mathbf{PC}_i$ is projected into $M$ depth maps $\{x_{i,m}\}_{m=1}^{M}$ for CLIP-based inference. However, not all projections contribute equally occluded or sparsely sampled views often yield ambiguous predictions that distort feature aggregation. To ensure that only structurally reliable projections influence adaptation, CLIPoint3D employs an \textit{entropy-guided view selection} mechanism that filters views based on prediction uncertainty.

For each view $x_{i,m}$, the posterior probability $p(y_k \mid x_{i,m})$ from Eq.~\eqref{eq:clip_prob} is used to compute predictive entropy:
\begin{equation}
\mathrm{H}_{i,m}
=
- \sum_{k=1}^{K}
p(y_k \mid x_{i,m}) \,
\log p(y_k \mid x_{i,m}),
\label{eq:entropy}
\end{equation}
where low $\mathrm{H}_{i,m}$ denotes high model confidence and geometric reliability. 
Views satisfying $\mathrm{H}_{i,m} \le \tau_\rho$, where $\tau_\rho$ is the $\rho$-th percentile of entropy values across all $M$ views of $\mathbf{PC}_i$ (we use $\rho = 0.5$), form the confident subset $\mathcal{M}_i^{*} = \{\, m \mid \mathrm{H}_{i,m} \le \tau_\rho \,\}$.
The class probability for the full point cloud is then aggregated over this selected subset:
\begin{equation}
p(y \mid \mathbf{PC}_i)
=
\frac{1}{|\mathcal{M}_i^{*}|}
\sum_{m \in \mathcal{M}_i^{*}}
p(y \mid x_{i,m}).
\label{eq:final_prob}
\end{equation}

We use $p_S$ and $p_T$ while considering the source and target samples, respectively.
This filtering strategy provides a self-adaptive reliability prior, retaining diverse yet confident projections while suppressing noisy or redundant ones. Unlike uniform pooling~\cite{pointclip}, it introduces no additional parameters and naturally adjusts to domain-dependent uncertainty (see Table~\ref{tab:view_selection}). 
Used in both training and inference, it complements PEFT by supplying uncertainty-aware evidence selection, yielding stable multi-view generalization.

\subsubsection{Domain Alignment Strategies}
\label{sec:uncertainty_alignment}

The final stage of {CLIPoint3D} aligns source and target distributions in CLIP's multimodal space while explicitly accounting for prediction uncertainty. Prior modules improve view reliability and semantic grounding, yet residual misalignment persists, mainly because conventional adversarial~\cite{dann} or MMD-based~\cite{mmd} methods treat all samples equally, allowing low-confidence target embeddings to dominate optimization.  
To address this, we propose an \textit{Uncertainty-Aware Optimal Transport (UA-OT)} framework that introduces {two complementary novelties}:  
(i)~\textbf{entropy-weighted class prototypes} that perform sample-wise confidence filtering at the class level, and  
(ii)~\textbf{entropy-regularized OT} that enforces global distribution matching while suppressing noisy couplings.  
Together, these mechanisms deliver confidence-calibrated alignment unavailable in prior 3D/2D DA methods.

\noindent \textbf{Entropy-weighted prototype alignment.}
Let $\mathbf{v}_i^{S}$ and $\mathbf{v}_j^{T}$ denote the confident-view aggregated embeddings of source and target clouds (Sec.~\ref{sec:entropy_view}).  
Each source cloud is assigned an entropy-based reliability weight,
\begin{equation}
w_i^{S}
=
1 -
\frac{\mathrm{H}\!\left(p_{S}(y \mid \mathbf{PC}_i^{S})\right)}
{\log K},
\end{equation}
allowing uncertain samples to be automatically down-weighted.  
The resulting class-specific prototype is
\begin{equation}
\label{eq:proto_multi}
\mathbf{U}_{c} =
\frac{
\sum_{i:\,y_i^{S}=c}
w_{i}^{S}\,\mathbf{v}_{i}^{S}
}{
\sum_{i:\,y_i^{S}=c} 
w_{i}^{S}
}.
\end{equation}

Target clouds receive analogous weights and pseudo-labels:
\begin{equation}
\begin{aligned}
w_{j}^{T} &= 
1 -
\frac{\mathrm{H}\!\left(p_{T}(y \mid \mathbf{PC}_j^{T})\right)}{\log K},
\\[4pt]
\hat{y}_{j} &= 
\arg\max_{c}\, p_{T}(y=c \mid \mathbf{PC}_j^{T}).
\end{aligned}
\end{equation}
Prototype alignment is enforced via
\begin{equation}
\label{eq:proto_loss_multi}
\mathbf{L}_{\mathrm{proto}} =
-\sum_{j}
\, w_{j}^{T}\,
\log
\frac{
\exp\!\left(
\cos(\mathbf{v}_{j}^{T}, \mathbf{U}_{\hat{y}_j})/\tau
\right)
}{
\sum_{c'}
\exp\!\left(
\cos(\mathbf{v}_{j}^{T}, \mathbf{U}_{c'})/\tau
\right)
}.
\end{equation}
This \emph{uncertainty-weighted class coupling} is a key novelty: high-confidence target clouds drive semantic alignment, while unreliable ones contribute minimally.

\noindent \textbf{Entropy-regularized optimal transport.}
While prototype alignment promotes class-conditional consistency, global distribution mismatch can persist.  
We therefore apply a second novelty, \textbf{entropy-regularized OT over cloud-level embeddings}, which provides smooth, noise-tolerant domain matching.  
Let
\[
C_{ij} = \|\mathbf{v}_{i}^{S} - \mathbf{v}_{j}^{T}\|_2^{2}
\]
be the transport cost and  
$\pi \in \Pi(\mathcal{P}_S,\mathcal{P}_T)$ a feasible plan.  
The UA-OT loss is
\begin{equation}
\label{eq:ot_multi}
\mathbf{L}_{\mathrm{OT}}
=
\min_{\pi}
\Big(
\langle C,\pi\rangle
-
\varepsilon H(\pi)
\Big),
\end{equation}
where the entropy term $H(\pi)$ avoids overly sharp couplings and stabilizes the alignment under prediction noise.

\noindent \textbf{Auxiliary calibration loss.}
To further support stable prototype and OT coupling, we minimize the prediction entropy of both domains:

\begin{equation}
\label{eq:conf_combined}
\mathbf{L}_{\mathrm{conf}} =
\sum_{i}
\mathrm{H}\!\left(p_{S}(y \mid \mathbf{PC}_i^{S})\right) +
\frac{1}{N_T}
\sum_{j}
\mathrm{H}\!\left(p_{T}(y \mid \mathbf{PC}_j^{T})\right)
\end{equation}

The first term yields cleaner source prototypes, while the second encourages compact target clusters that reinforce Eq.~\ref{eq:proto_loss_multi}.  
Together, {entropy-weighted prototypes}, {entropy-regularized OT}, and unified calibration form a robust, confidence-driven alignment mechanism.

\subsection{Overall Training and Inference}
\label{sec:training_inference}

\noindent\textbf{Training.} The full optimization objective of {CLIPoint3D} integrates supervised learning, geometric regularization, and uncertainty-aware alignment in a unified framework. The total loss is
\begin{equation}
\label{eq:loss_total}
\mathbf{L}_{\mathrm{total}}
=
\mathbf{L}_{\mathrm{ce}}
+
\alpha\!\left(
\mathbf{L}_{\mathrm{ortho}}
+
\mathbf{L}_{\mathrm{proto}}
+
\mathbf{L}_{\mathrm{OT}}
+
\mathbf{L}_{\mathrm{conf}}
\right),
\end{equation}
where $\mathbf{L}_{\mathrm{ce}}$ is the supervised cross-entropy on source data $\mathcal{D}_S$.
The geometric consistency term $\mathbf{L}_{\mathrm{ortho}}$~\cite{pointnet} regularizes the 3D encoder $\mathcal{E}_{3D}$ by enforcing local feature decorrelation, for both the domains:
\begin{equation}
\mathbf{L}_{\mathrm{ortho}}
=
\big\|\mathbf{I}_{3D}^\top \mathbf{I}_{3D}-\mathbb{I}\big\|_2^2,
\end{equation}
where $\mathbb{I}$ is the identity matrix.

\noindent\textbf{Inference.} At test time, each target cloud $\mathbf{PC}_j^{T}$ is projected into multiple views, and predictions are aggregated using the entropy-guided selection rule (Eq.~\ref{eq:final_prob}), yielding a robust multi-view decision.

\subsection{Generalization Bound}
\label{sec:gen_bound}

Following classical DA theory~\cite{ben2006analysis,redko2017theoretical},
for any hypothesis $h\!\in\!\mathcal{H}$ with bounded loss $\ell\!\in\![0,1]$, the source risk is
\begin{equation}
\mathcal{R}_{S}(h)=
\mathbb{E}_{(\mathbf{x},y)\sim\mathcal{P}_{S}}
[\ell(h(\mathbf{x}),y)].
\end{equation}
The corresponding target risk is upper-bounded by
\begin{equation}
\mathcal{R}_{T}(h)
\le
\mathcal{R}_{S}(h)
+\tfrac12 d_{\mathcal{H}\Delta\mathcal{H}}(\mathcal{P}_{S},\mathcal{P}_{T})
+\lambda^{*},
\label{eq:da_bound_mid}
\end{equation}
where $d_{\mathcal{H}\Delta\mathcal{H}}$ measures distributional discrepancy and 
$\lambda^{*}$ is the joint optimal risk over $\mathcal{H}$.

In CLIPoint3D,  
the entropy-regularized OT loss $W_{\!\varepsilon}(\mathcal{P}_{S},\mathcal{P}_{T})$
serves as a smooth surrogate for $d_{\mathcal{H}\Delta\mathcal{H}}$,
providing a stable measure of global domain shift.
Complementarily, the uncertainty-weighted prototype alignment suppresses noisy features
and encourages class-conditional consistency, thereby reducing the discrepancy contributing to $\lambda^{*}$.
Let $\mathbf{U}_{c}^{S}$ and $\mathbf{U}_{c}^{T}$ denote entropy-weighted
source and pseudo-labeled target prototypes;
their agreement enforces tight semantic coupling across domains.

Under these relaxations, the resulting surrogate bound becomes
\begin{equation}
\mathcal{R}_{T}(h)
\le
\mathcal{R}_{S}(h)
+\tfrac12 W_{\!\varepsilon}(\mathcal{P}_{S},\mathcal{P}_{T})
+\beta\sum_{c=1}^{K}\|\mathbf{U}_{c}^{S}-\mathbf{U}_{c}^{T}\|_{2}^{2},
\label{eq:gen_bound_final_mid}
\end{equation}
where $\beta$ trades off global (OT) and semantic (prototype) alignment.
Entropy-guided view selection lowers $\mathcal{R}_{S}(h)$
by filtering uncertain projections.
Together, these components yield a tighter, uncertainty-aware bound
and support reliable transfer of CLIP’s 2D priors into the 3D setting. See \texttt{Supplementary} for further discussions.

\section{Experimental Evaluation}
\label{sec:experiments}
\noindent\textbf{Datasets:}
We evaluate our proposed method on two domain adaptation benchmarks: PointDA-10 \cite{pointdan} and GraspNetPC-10 \cite{gai}. The PointDA-10 benchmark consists of three widely used PC datasets: ModelNet \cite{modelnet}, ShapeNet \cite{shapenet}, and ScanNet \cite{scannet}. In contrast, GraspNetPC-10, derived from GraspNet \cite{graspnet} by \cite{gai}, includes three distinct domains: Synthetic, Kinect, and RealSense. Both benchmarks share the same 10 object categories across all domains. Further details are provided in the \texttt{Supplementary}.

\noindent\textbf{Implementation Details and Evaluation Metric:} In our experiments, we use a single A100 GPU of 80 GB and employ the frozen ViT-B/16 variant of the CLIP backbone, PointNet~\cite{pointnet} as the 3D encoder, and GPT-5~\cite{gpt5} as the LLM.  
Each multi-head cross-attention block comprises four attention heads, followed by layer normalization and an FFN with a two-layer bottleneck structure (\texttt{Linear-GeLU-Linear}).  
We use $M=10$ projected depth maps for each point cloud sample.  
The learnable query vector $\mathbf{q}$ has a length of 4 and dimensionality of 512.  
For parameter-efficient fine-tuning, we adopt LoRA~\cite{lora} with a rank of 16 and dropout rate of 0.1.  
In Eq.~\ref{eq:loss_total}, the loss balancing coefficient is set to $\alpha=1$.  
Training is performed with a batch size of 32, 64 shots per class, and for a total of 50 epochs.  
The learning rate is initialized at 0.002 with a decay rate of $1\times10^{-5}$, momentum of 0.9, and the SGD \cite{sgd} optimizer. We report the classification performance in the target domain as the evaluation metric. Reported results are averaged over three runs.

\subsection{Comparisons to the Literature}
\label{sec:comparisons}

\begin{table}[t]
    \centering
    \caption{\textbf{Domain adaptation performance on the PointDA-10 benchmark.} 
    M: ModelNet, S: ShapeNet, S$^{*}$: ScanNet; $\rightarrow$ indicates the adaptation direction. Best results and second-best results are reported in bold and underlined, respectively.}
    \scalebox{0.68}{
    \begin{tabular}{l|cccccc|c}
    \toprule
        \textbf{Methods} & \textbf{M$\rightarrow$S}  & \textbf{M$\rightarrow$S$^{*}$} & \textbf{S$\rightarrow$M} & \textbf{S$\rightarrow$S$^{*}$} & \textbf{S$^{*}\rightarrow$M} & \textbf{S$^{*}\rightarrow$S} & \textbf{Avg} \\
    \midrule
        DANN \cite{dann} &74.8 &42.1 &57.5 &50.9 &43.7 &71.6 &56.8 \\
        PointDAN \cite{pointdan} &83.9 &44.8 &63.3 &45.7 &43.6 &56.4 &56.3 \\
        RS \cite{rs} &79.9 &46.7 &75.2 &51.4 &71.8 &71.2 &66.0 \\
        DAE-Global \cite{daeglobal}	&83.5	&42.6	&74.8	&45.5	&64.9	&67.3	&63.1 \\
        DefRec \cite{defrec} &82.7 &43.9 &\underline{79.8} &48.0 &66.0 &67.4 &64.6 \\
        DefRec + PCM \cite{defrec} &83.3 &53.5 &78.5 &53.2 &73.7 &75.5 &69.6 \\
        GAST \cite{gast} &83.9 &\underline{56.7} &76.4 &55.0 &73.4 &72.2 &69.5 \\
        GAI \cite{gai} &\textbf{85.8} &55.3 &77.2 &55.4 &73.8 &72.4 &70.0 \\
        MLSP \cite{mlsp} &83.7 &55.4 &77.1 &\underline{55.6} &\underline{78.2} &76.1 &71.0 \\
        3DeNet \cite{3denet} &84.5 &\textbf{57.1} &78.8 &\textbf{57.2} &77.5 &\underline{78.1} &\underline{72.2} \\
        \midrule
        ZS-CLIP \cite{clip}  & 46.1& 17.0& 52.0& 17.0& 52.0& 46.1 & 38.4\\
        PointCLIP \cite{pointclip} &50.8 &20.9 & 50.1& 20.9& 50.1& 50.8& 40.6\\
        PointCLIPv2 \cite{pointclipv2} &38.8 & 19.5& 71.6& 19.5& 71.6& 38.8& 43.3\\
        \midrule

        \cellcolor[gray]{0.93}\textbf{CLIPoint3D-T} &\cellcolor[gray]{0.93}74.4 &\cellcolor[gray]{0.93}9.5 &\cellcolor[gray]{0.93}86.0 &\cellcolor[gray]{0.93}24.1 &\cellcolor[gray]{0.93}50.5 &\cellcolor[gray]{0.93}59.8 &\cellcolor[gray]{0.93}50.7 \\
        
        \cellcolor[gray]{0.93}\textbf{CLIPoint3D-V} &\cellcolor[gray]{0.93}\underline{84.6} &\cellcolor[gray]{0.93}53.5 &\cellcolor[gray]{0.93}\textbf{91.6} &\cellcolor[gray]{0.93}55.3 &\cellcolor[gray]{0.93}\textbf{87.9} &\cellcolor[gray]{0.93}81.3 &\cellcolor[gray]{0.93}\textbf{75.7} \\

        \cellcolor[gray]{0.93}\textbf{CLIPoint3D-B} &\cellcolor[gray]{0.93}81.5 &\cellcolor[gray]{0.93}51.9 &\cellcolor[gray]{0.93}90.3 &\cellcolor[gray]{0.93}46.6 &\cellcolor[gray]{0.93}85.2 &\cellcolor[gray]{0.93}\textbf{85.8} &\cellcolor[gray]{0.93}73.6 \\
        
        \midrule
        \cellcolor{blue!15} Improvement &\cellcolor{blue!15}\textcolor{darkred}{-\textbf{1.2}} &\cellcolor{blue!15}\textcolor{darkred}{-\textbf{3.6}} &\cellcolor{blue!15}\textcolor{darkgreen}{+\textbf{11.8}} &\cellcolor{blue!15}\textcolor{darkred}{-\textbf{1.9}} &\cellcolor{blue!15}\textcolor{darkgreen}{+\textbf{9.7}} &\cellcolor{blue!15}\textcolor{darkgreen}{+\textbf{7.7}} &\cellcolor{blue!15}\textcolor{darkgreen}{+\textbf{3.5}} \\
    \bottomrule
    \end{tabular}}
    \label{tab:pointda10}
\vspace{-0.1cm}  
\end{table}

 Table~\ref{tab:pointda10} and \ref{tab:graspnetpc10} detail the comparative analysis of CLIPoint3D against leading conventional encoder-based and CLIP-based approaches.  
Suffixes `T', `V', and `B' indicate LoRA fine-tuning on CLIP's text, vision, and both encoders, respectively.

\noindent\textbf{Results on PointDA-10.}  
Table~\ref{tab:pointda10} shows that zero-shot CLIP variants (PointCLIP, PointCLIPv2, ZS-CLIP) underperform encoder-based methods, particularly on the challenging \textit{ScanNet} domain.  
While encoder-based models capture geometric cues more effectively, they still lag behind our {CLIPoint3D}.  
Across all source-target pairs, {CLIPoint3D} achieves the best average accuracy, surpassing prior approaches by at least \textbf{3.5\%}.  
Although it shows minor drops in certain cases (e.g., ScanNet adaptation), it yields substantial gains in synthetic domains (\textit{ModelNet}, \textit{ShapeNet}), ranking second for ModelNet$\rightarrow$ShapeNet.

\noindent\textbf{Results on GraspNetPC-10.}  
Table~\ref{tab:graspnetpc10} highlights that encoder-based methods exhibit unstable synthetic-to-real performance, especially under \textit{Kinect} and \textit{Realsense} sensors.  
In contrast, {CLIPoint3D} achieves consistent improvements with an average margin of \textbf{16.4\%} across all adaptation directions.  
It maintains strong generalization under both synthetic-to-real and real-to-real shifts.  
Zero-shot CLIP baselines remain weak in transferability, underscoring their limited 3D adaptability.  
Overall, {CLIPoint3D} delivers the most balanced and robust results, effectively bridging vision-language pretraining with 3D adaptation.

\begin{table}[t]
    \centering
    \caption{\textbf{Domain adaptation performance on the GraspNetPC-10 benchmark.} Syn.: Synthetic domain, Kin.: Kinect domain, RS.: Realsense domain; $\rightarrow$ indicates the adaptation direction. }
    \scalebox{0.68}{
    \begin{tabular}{l|cccc|c}
    \toprule
        \textbf{Methods} & \textbf{Syn.$\rightarrow$Kin.}  & \textbf{Syn$\rightarrow$RS.} & \textbf{Kin.$\rightarrow$RS.} & \textbf{RS.$\rightarrow$Kin.} &\textbf{Avg} \\
    \midrule
        DANN \cite{dann} &78.6 &70.3 &46.1 &67.9 &65.7 \\
        PointDAN \cite{pointdan} &77.0 &72.5 &65.9 &82.3 &74.4 \\
        RS \cite{rs} &67.3 &58.6 &55.7 &69.6 &62.8 \\
        DefRec + PCM \cite{defrec} &80.7 &70.5 &65.1 &77.7 &73.5 \\
        GAST \cite{gast} &69.8 &61.3 &58.7 &70.6 &65.1 \\
        GAI \cite{gai} &\underline{81.2} &\underline{73.1} &\underline{66.4} &\underline{82.6} &\underline{75.8} \\
        \midrule
        ZS-CLIP \cite{clip} & 20.0& 14.8& 14.8& 20.0& 17.4\\
        PointCLIP \cite{pointclip} &30.7 &24.3 &24.3 & 30.7& 27.5\\
        PointCLIPv2 \cite{pointclipv2} & 30.3& 22.8& 22.8& 30.3& 26.6\\
        \midrule
        \cellcolor[gray]{0.93}\textbf{CLIPoint3D-T} &\cellcolor[gray]{0.93}87.6 &\cellcolor[gray]{0.93}71.6 &\cellcolor[gray]{0.93}74.2 &\cellcolor[gray]{0.93}82.3 &\cellcolor[gray]{0.93}78.9 \\

        \cellcolor[gray]{0.93}\textbf{CLIPoint3D-V} &\cellcolor[gray]{0.93}95.0 &\cellcolor[gray]{0.93}85.0 &\cellcolor[gray]{0.93}\textbf{88.4} &\cellcolor[gray]{0.93}94.3 &\cellcolor[gray]{0.93}90.7 \\

        \cellcolor[gray]{0.93}\textbf{CLIPoint3D-B} &\cellcolor[gray]{0.93}\textbf{96.5} &\cellcolor[gray]{0.93}\textbf{89.3} &\cellcolor[gray]{0.93}86.8 &\cellcolor[gray]{0.93}\textbf{96.2} &\cellcolor[gray]{0.93}\textbf{92.2} \\

        \midrule
        \cellcolor{blue!15}Improvement &\cellcolor{blue!15}\textcolor{darkgreen}{+\textbf{15.3}} &\cellcolor{blue!15}\textcolor{darkgreen}{+\textbf{16.2}} &\cellcolor{blue!15}\textcolor{darkgreen}{+\textbf{22.0}} &\cellcolor{blue!15}\textcolor{darkgreen}{+\textbf{13.6}} &\cellcolor{blue!15}\textcolor{darkgreen}{+\textbf{16.4}} \\

    \bottomrule
    \end{tabular}}
    \label{tab:graspnetpc10}
\vspace{-0.1cm}  
\end{table}

\section{Ablation Studies}
\label{sec:ablations}

\noindent\textbf{(i) Impact of PEFT methods.}  
Table~\ref{tab:peft} compares PEFT adaptation strategies such as LayerNorm tuning~\cite{layernorm}, BitFit~\cite{bitfit}, and LoRA~\cite{lora,cliplora,2sfs,fedmvp} within CLIPoint3D.  
Among the standalone methods, LoRA achieves the highest accuracy (\textit{90.5\%}), especially when jointly applied to both encoders, while BitFit and LayerNorm tuning offer limited gains. This shows that low-rank adaptation captures domain-specific cues more effectively than simple bias or normalization tuning. Combining LoRA with our proposed prompting method improves results ( \textit{92.2\%}), confirming their complementarity in refining task-specific subspaces.

\begin{table}[t]
    \centering
    \caption{\textbf{Ablation study of PEFT methods.} Here, `PT' refers to our proposed knowledge-driven prompt tuning strategy. The results reported are the average performances on GraspNetPC-10.}
    \vspace{-0.2cm}
    \scalebox{0.68}{
    \begin{tabular}{l|ccc|c|ccc}
        \toprule
        & \multicolumn{3}{c|}{\textbf{PEFT}} & \multicolumn{1}{c|}{\textbf{PT}} & \multicolumn{3}{c}{\textbf{PEFT + PT}} \\
        \cmidrule(lr){2-4} \cmidrule(lr){5-5} \cmidrule(lr){6-8}
        \textbf{Method} & Text & Vision & Both & - & Text & Vision & Both \\
        \midrule
        \cellcolor[gray]{0.93}LoRA \cite{lora}     &\cellcolor[gray]{0.93}80.3 &\cellcolor[gray]{0.93}89.2 &\cellcolor[gray]{0.93}90.5 & ~ &\cellcolor[gray]{0.93}78.9 &\cellcolor[gray]{0.93}\underline{90.7} &\cellcolor[gray]{0.93}\textbf{92.2} \\
        LayerNorm \cite{layernorm} &75.4 &\textbf{84.0} &76.6 & 73.7 &75.0 &\underline{79.5} &78.8 \\
        BitFit \cite{bitfit}   &80.1 &\underline{87.8} &81.1 & ~ &79.7 &\textbf{88.5} &81.5 \\
        \bottomrule
    \end{tabular}}
    \label{tab:peft}
\end{table}

\noindent\textbf{(ii) Sensitivity to loss functions.}  
Table~\ref{tab:loss} reports the contribution of each loss term.  
The baseline with only $\mathcal{L}_{\text{ce}}$, as well as combinedly with $\mathcal{L}_{\text{ortho}}$ and $\mathcal{L}_{\text{conf}}$ showcase the lack of domain alignment and yield lower accuracy.  
Adding $\mathcal{L}_{\text{proto}}$ improves semantic consistency, while $\mathcal{L}_{\text{OT}}$ substantially reduces domain discrepancy. 
Optimizing all terms jointly achieves peak accuracy, \textit{75.7\%} (PointDA-10) and \textit{92.2\%} (GraspNetPC-10), demonstrating their roles.

\begin{table}[t]
    \centering
    \caption{\textbf{Ablation of different loss components.} The results reported are the average performance for benchmarks.}
    \scalebox{0.68}{
    \begin{tabular}{ccccc|c|c}
    \toprule
        $\mathbf{L}_{\mathrm{ce}}$ &    
        $\mathbf{L}_{\mathrm{ortho}}$ &
        $\mathbf{L}_{\mathrm{proto}}$  & $\mathbf{L}_{\mathrm{OT}}$ & $\mathbf{L}_{\mathrm{conf}}$ 
        & \textbf{PointDA-10} & \textbf{GraspNetPC-10} \\
    \midrule
    \cmark & \xmark & \xmark & \xmark & \xmark &49.8  &64.3 \\

    \cmark & \cmark & \xmark & \xmark & \xmark &58.6  &74.9 \\

    \cmark & \cmark & \cmark & \xmark & \xmark & 64.5 & 80.3 \\
    \cmark & \cmark & \xmark & \cmark & \xmark & 70.4 & 85.0 \\
    \cmark & \cmark & \xmark & \xmark & \cmark & 71.3 & 81.9 \\
    \cmark & \cmark  & \cmark & \cmark & \xmark & 71.0 & 85.8 \\
    \cmark & \cmark  & \xmark & \cmark & \cmark & 73.9 & 86.3 \\
    \cmark & \cmark  & \cmark & \xmark & \cmark &72.3 & 84.3 \\
    \cellcolor[gray]{0.93}\cmark 
    & \cellcolor[gray]{0.93}\cmark 
    &\cellcolor[gray]{0.93}\cmark 
    &\cellcolor[gray]{0.93}\cmark 
    &\cellcolor[gray]{0.93}\cmark 
    &\cellcolor[gray]{0.93}\textbf{75.7} 
    &\cellcolor[gray]{0.93}\textbf{92.2} \\
        
    \bottomrule
    \end{tabular}}
    \label{tab:loss}
    \vspace{-0.1cm}  
\end{table}

\noindent\textbf{(iii) Ablation on prompting strategies.}  
Table~\ref{tab:prompting} compares different prompting configurations.  
Using only textual prompts $\mathbf{P}_t(\mathbf{q})$ or visual prompts $\mathbf{P}_v(\mathbf{q})$ yields moderate gains, with $\mathbf{P}_v$ performing slightly better due to geometric cues.  
Naive multimodal concatenation (MaPLe~\cite{maple}) fails to exploit cross-modal complementarity effectively, but achieves better performance than unimodal prompting.  
In contrast, our LLM-guided textual prompts $\mathbf{P}_t(\mathbf{T}^{\text{llm}}, \mathbf{q})$ and 3D-conditioned visual prompts $\mathbf{P}_v(\mathbf{I}_{3D}, \mathbf{q})$ jointly achieve the highest accuracy (\textit{75.7\%}), confirming that semantic grounding and geometric awareness act synergistically to enable robust multimodal adaptation.

\begin{table}[t]
    \centering
    \caption{\textbf{Analysis of our prompting strategy in PointDA-10 benchmark.} Explanations of the notations below are given in Section \ref{sec:methodology}. M: ModelNet, S: ShapeNet, S$^{*}$: ScanNet.}
    \scalebox{0.62}{
    \begin{tabular}{l|cccccc|c}
    \toprule
        \textbf{Strategy} & \textbf{M$\rightarrow$S}  & \textbf{M$\rightarrow$S$^{*}$} & \textbf{S$\rightarrow$M} & \textbf{S$\rightarrow$S$^{*}$} & \textbf{S$^{*}\rightarrow$M} & \textbf{S$^{*}\rightarrow$S} & \textbf{Avg} \\
    \midrule

    $\mathbf{P}_{t}(\mathbf{q})$ only &80.1 &45.6 &84.4 &44.3 &81.5 &72.3 &68.0 \\

    $\mathbf{P}_{v}(\mathbf{q})$ only &82.1 &48.9 &85.7 &48.1 &82.1 &71.7 &69.8 \\

    $\mathbf{P}_{t}(\mathbf{q})$ + $\mathbf{P}_{v}(\mathbf{q})$ &\textbf{85.4} &50.8 &90.8 &48.4 &87.5 &71.3 &72.4 \\

    $\mathbf{P}_{t}(\mathbf{q})$ +  $\mathbf{P}_{v}(\mathbf{I}_{3D}, \mathbf{q})$ &83.2 &49.5 &91.2 &48.9 &83.9 &71.1 &71.3 \\

    $\mathbf{P}_{t}(\mathbf{T}^{\text{llm}}, \mathbf{q})$ +  $\mathbf{P}_{v}(\mathbf{q})$ &81.4 &52.3 &91.2 &49.2 &84.5 &\textbf{89.2} &74.6 \\
    \midrule

    \cellcolor[gray]{0.93}$\mathbf{P}_{t}(\mathbf{T}^{\text{llm}}, \mathbf{q})$ +  $\mathbf{P}_{v}(\mathbf{I}_{3D}, \mathbf{q})$ &\cellcolor[gray]{0.93}84.6 
    &\cellcolor[gray]{0.93}\textbf{53.5} 
    &\cellcolor[gray]{0.93}\textbf{91.6} &\cellcolor[gray]{0.93}\textbf{55.3} 
    &\cellcolor[gray]{0.93}\textbf{87.9} 
    &\cellcolor[gray]{0.93}81.3 
    &\cellcolor[gray]{0.93}\textbf{75.7} \\
    
    \bottomrule
    \end{tabular}}
    \label{tab:prompting}
\vspace{-0.1cm}  
\end{table}

\begin{figure}[h]
    \centering
    \begin{subfigure}[t]{0.48\linewidth}
        \centering
        \caption{\centering \textbf{Few-Shots Comparison}}
        \includegraphics[width=\linewidth]{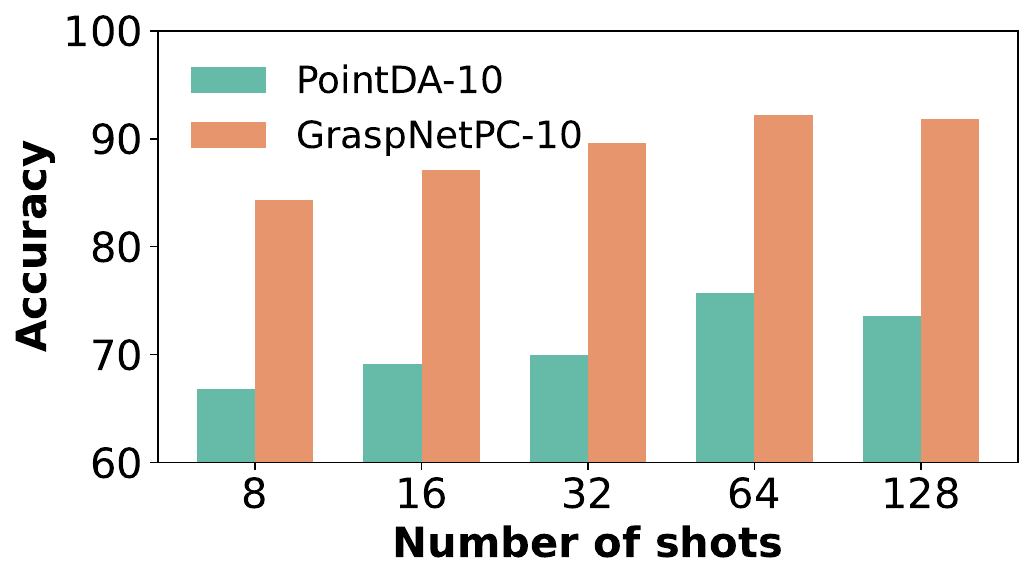}
        \label{fig:shots}
    \end{subfigure}
    \hfill
    \begin{subfigure}[t]{0.48\linewidth}
        \centering
        \caption{\centering \textbf{Multi-View Sensitivity}}
        \includegraphics[width=\linewidth]{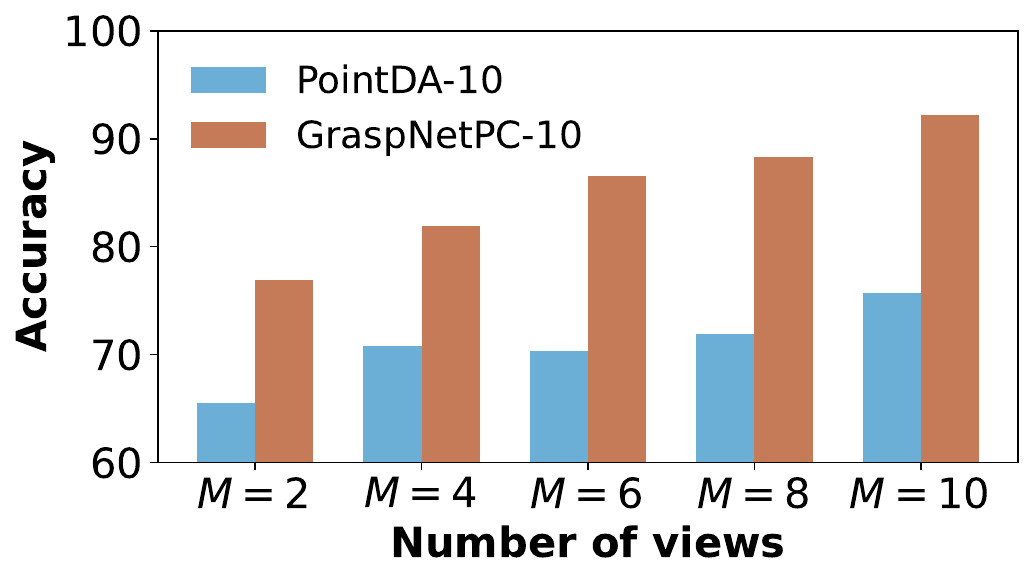}
        \label{fig:views}
    \end{subfigure}
    \vspace{-0.6cm}
    \caption{(a) Effect of the \textbf{number of labeled samples} in $\mathcal{D}_s$ during training. (b) \textbf{Effect of projected views.} Accuracy variation with projection count $M$.}
    \label{fig:shotview}
\end{figure}

\noindent\textbf{(iv) Effects of number of shots.}  
Figure~\ref{fig:shots} shows performance under varying the supervision level in $\mathcal{D}_s$.  
Accuracy rises sharply from as the training set size in $\mathcal{D}_s$ changes from $8$ to $64$, peaking at \textit{75.7\%} and \textit{92.2\%} on PointDA-10 and GraspNetPC-10, respectively, after which gains majorly saturate despite increasing the training data.

\noindent\textbf{(v) Influence of projected views.}  
As shown in Figure ~\ref{fig:views}, increasing the number of 2D projections $M$ enhances performance by enriching multi-view cues. However, we achieve the maximum accuracy at $M{=}10$, where additional views are found to add redundancy with minimal gain.

\noindent\textbf{(vi) Effect of view selection.}  
Table~\ref{tab:view_selection} analyzes different view aggregation strategies. Averaging logits performs best, while weighted average or max-similarity driven selection slightly degrades results. Random selection causes a large drop, confirming the value of informed sampling.  
Our entropy-guided approach yields the top accuracy, prioritizing dominated views for stable multi-view representation.

\noindent\textbf{(vii) Computational complexity.}  
Table~\ref{tab:complexity} compares model efficiency.  
While DANN and DefRec+PCM are lightweight but less accurate, large models like GAST (160M+ params) offer no advantage. CLIPoint3D achieves the best trade-off, having only 9-11M trainable parameters with SOTA performance on both benchmarks.

\begin{table}[t]
    \centering
    \caption{\textbf{Ablation of view selection strategies on GraspNetPC-10.} 
`Avg.', `W. avg.', and `Max Sim' denote uniform averaging, weighted averaging, and maximum-similarity-driven logit selection. `Random' selects a random single view, while `Entropy-guided' is our proposed uncertainty-based selection scheme.}

    \vspace{-0.1cm}
    \scalebox{0.65}{
    \begin{tabular}{l|ccccc}
    \toprule
        \textbf{Method} &\textbf{Avg.} &\textbf{W. Avg}	&\textbf{Random} &\textbf{Max Sim.} &\textbf{Entropy-guided} \\
    \midrule

    CLIPoint3D-T &\textbf{82.5}	&77.7	&51.4	&\underline{80.1} &\cellcolor[gray]{0.93}78.9 \\
    CLIPoint3D-V &\textbf{91.5}	&86.5	&71.5	&86.3 &\cellcolor[gray]{0.93}\underline{90.7} \\
    CLIPoint3D-B &\underline{92.0}	&91.1	&70.9   &85.4	&\cellcolor[gray]{0.93}\textbf{92.2} \\

    \bottomrule
    \end{tabular}}
    \label{tab:view_selection}
\end{table}

\begin{table}[t]
    \centering
    \caption{\textbf{Trade-off between computational complexity and model performances.} Trainable parameters and accuracy are reported in millions (M) and \%, respectively.}
    \vspace{-0.2cm}
    \scalebox{0.65}{
    \begin{tabular}{l|ccccc}
    \toprule
        \textbf{Methods} & \textbf{DANN} &\textbf{PointDAN} & \textbf{DAE-Global} 
        & \textbf{DefRec+PCM} & \textbf{GAST} \\
    \midrule

    Train params. (M) &\textbf{2.50} &11.26 &12.60 &\textbf{2.50} &161.09 \\
    PointDA-10 (\%) &56.8 &56.3 &63.1 & 69.6  &69.5\\
    GraspNetPC-10 (\%) &65.7 &74.4 & - &73.5 &65.1 \\

    \midrule

    \textbf{Methods} & \textbf{GAI} &\textbf{MLSP} & \textbf{Ours-T} 
        & \textbf{Ours-V} & \textbf{Ours-B} \\
    \midrule

    Train params. (M) &22.02 &28.0 &\cellcolor[gray]{0.93}\underline{9.24} &\cellcolor[gray]{0.93}\underline{9.83} &\cellcolor[gray]{0.93}\underline{11.00} \\
    PointDA-10 (\%) &70.0 &\underline{71.0} &\cellcolor[gray]{0.93}50.7 &\cellcolor[gray]{0.93}\textbf{75.7} &\cellcolor[gray]{0.93}73.6 \\
    GraspNetPC-10 (\%) &\underline{75.8} & - &\cellcolor[gray]{0.93}78.9 &\cellcolor[gray]{0.93}90.7 &\cellcolor[gray]{0.93}\textbf{92.2} \\

    \bottomrule
    \end{tabular}}
    \label{tab:complexity}
\vspace{-0.3cm}  
\end{table}

\begin{figure}[!htbp]
    \centering
    \includegraphics[width=1.05\columnwidth]{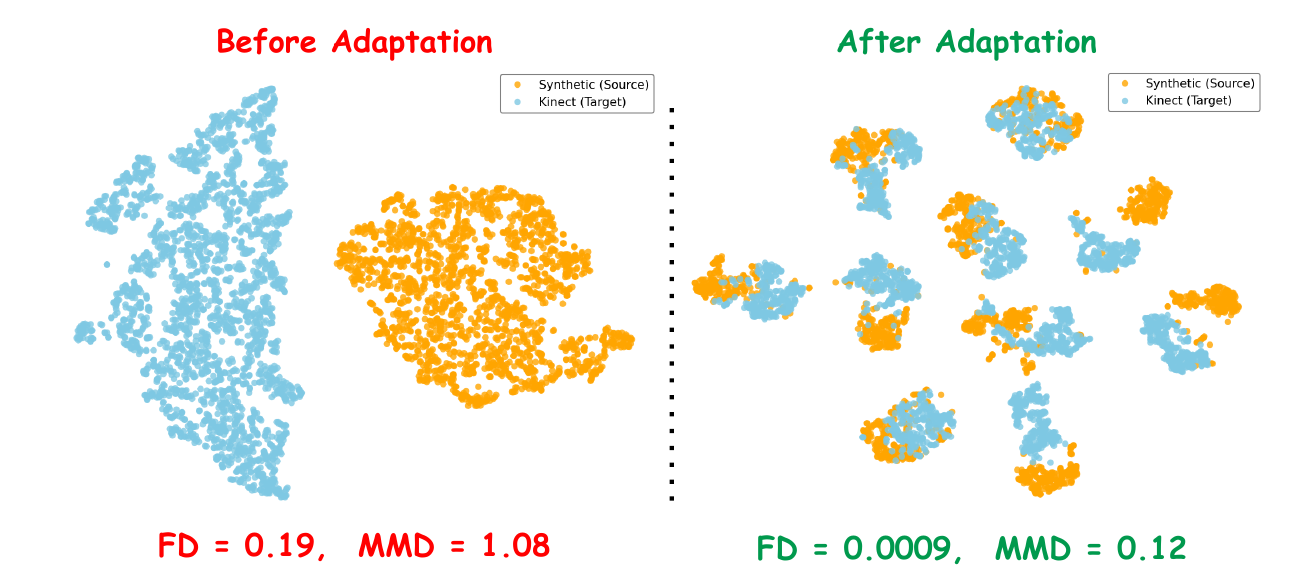}
    \vspace{-0.1cm}
    \caption{\textbf{t-SNE visualization} of CLIPoint3D's performance. Alignment between synthetic and real domains post-adaptation. FD and MMD quantify domain gap reduction.}
    \label{fig:tsne}
\end{figure}

\noindent\textbf{(viii) Qualitative results.}  
Figure~\ref{fig:tsne} visualizes source (synthetic) and target (Kinect) embeddings via t-SNE~\cite{tsne}.  
After adaptation, features form compact, overlapping clusters, indicating better alignment.  
Fréchet Distance \cite{frechet} drops from 0.19 to 0.0009 and MMD \cite{mmd} from 1.08 to 0.12, confirming improved cross-domain consistency.

\section{Conclusions}
\label{sec:conclusion}
In this work, we present CLIPoint3D, a framework for few-shot unsupervised 3D point cloud domain adaptation built on CLIP. By integrating knowledge-driven prompt tuning, PEFT, and entropy-guided view sampling, CLIPoint3D adapts VLMs to 3D without retraining large encoders, while optimal transport and uncertainty-aware prototype losses ensure robust domain alignment and class discriminability. Extensive results on PointDA-10 and GraspNetPC-10 show consistent gains over the state-of-the-art baselines by 3-16\% of improved performances.

\noindent \textbf{Acknowledgements.} This project has received funding from the European Union's EU Framework Programme for Research and Innovation under the HORIZON-MSCA-DN-2023 Grant Agreement N. 101169439. The work is also supported by the EU project ELIAS (no:101120237) and ELLIOT (no:101214398).

{
    \small
    \bibliographystyle{ieeenat_fullname}
    \bibliography{main}
}
\appendix
\setcounter{figure}{0}
\renewcommand{\thefigure}{A\arabic{figure}}

\setcounter{table}{0}
\renewcommand{\thetable}{A\arabic{table}}
\section{Supplementary Contents}
\begin{figure}[!htbp]
    \centering
    \includegraphics[width=1.05\columnwidth]{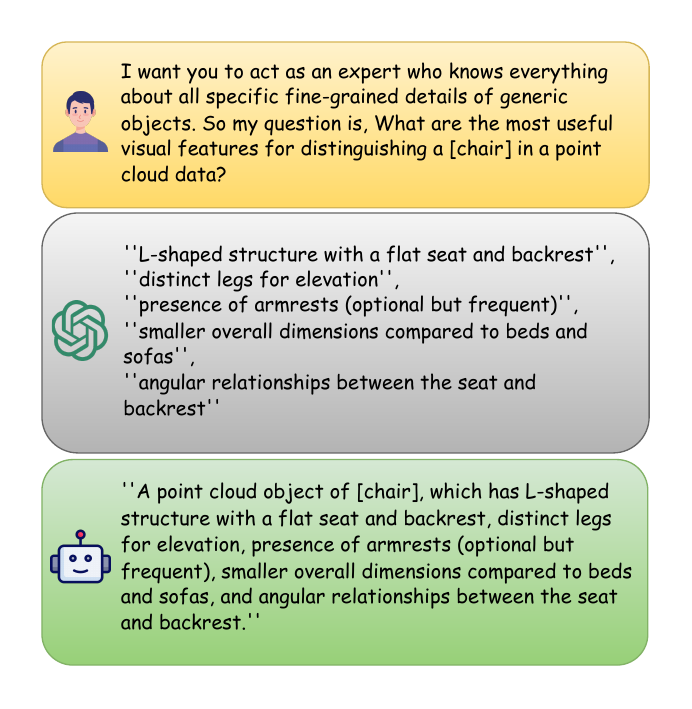}
    \caption{\textbf{LLM attributes generation.} To derive high-level 3D knowledge representations, we follow a three-stage pipeline. First (top box), we provide an instructional query prompt to a LLM (e.g. GPT-5 \cite{gpt5}). In response, the LLM produces detailed, geometry-aware visual descriptions (middle box). Finally (bottom box), we generate highly contextualized textual prompts (one caption per class) by combining a modality-specific prefix template with the LLM-generated attributes.}
    \label{fig:attributes}
\end{figure}

In this supplementary document, we present detailed information and further experimental results, including:

\begin{enumerate}
    \item \textbf{Dataset descriptions:} In Table \ref{tab:dataset}, we provide the total number of point cloud samples in the training and test splits of each domain of each datasets, though we do few-shot training in our proposed method.
    \item \textbf{LLM attributes generation:} In Fig. \ref{fig:attributes}, we show the pipeline of generation of high-level knowledge attributes using an LLM.
    \item \textbf{Pseudo-code of the CLIPoint3D algorithm:} In Algorithm \ref{pseudocode}, we provide the detailed procedure of our proposed method through a pseudo-code.
    \item \textbf{Analysis of the LoRA rank:} In Fig. \ref{fig:rank} we showcase the effect of the rank of LoRA metrices in CLIPoint3D on both datasets.
    \item \textbf{Conventional plug-in UDA methods in CLIP baselines:}  In Table \ref{tab:uda} we provide an analysis of training our CLIP-based zero-shot baselines using traditional UDA methods e.g.  DANN \cite{dann}, CDAN \cite{cdan} and SCDA \cite{scda}.
    
    \item \textbf{Effect of $\alpha$ hyperparameter:} In Fig. \ref{fig:alpha}, we showcase the importance of $\alpha$ hyperparameter used in the total loss function of Eq. 13.
    \item \textbf{Influence of the length of prompt $\mathbf{q}$:} In Fig. \ref{fig:length}, we show the effect of shared prompt length in CLIPoint3D.
    \item \textbf{Impact of CLIP variants:} In Table \ref{tab:backbone} we analyze the effect of CLIP backbones i.e. ViT-B/16, ViT-B/32 and ViT-L/14 in our proposed CLIPoint3D method.
    \item \textbf{Effect of various LLMs:} We also ablate how the attributes generated from different LLMs e.g. GPT-5 \cite{gpt5}, Llama-3.2-3B \cite{llama3}, Qwen2.5-14B \cite{qwen2}, Phi-4 \cite{phi4} etc can effect the performances of our method.
\end{enumerate}

\begin{table*}[htbp!]
    \centering
    \caption{Domain Generation dataset statistical details on class, training and test splits, prefix template.} 
     \centering
     \scalebox{0.88}{
     \begin{tabular}{c|c|c|c|c}
    \toprule     
        \textbf{Dataset} & \textbf{Domains} & \textbf{Common Classes} & \textbf{\# Samples} & \textbf{\# Training / Test} \\\\ 
        \midrule

         \multirow{3}{*}{PointDA-10 \cite{pointdan}} &ModelNet \cite{modelnet} &Bathtub, Bed, Bookshelf, &5039	&4183 / 856\\
         \cmidrule(lr){2-2}\cmidrule(lr){4-5}
         &ShapeNet \cite{shapenet} &Cabinet, Chair, Lamp, &19870	&17378 / 2492\\
         \cmidrule(lr){2-2}\cmidrule(lr){4-5}
         &ScanNet \cite{scannet} &Monitor, Plant, Sofa, Table &7879 &6110 / 1769\\
         \midrule
         
         \multirow{3}{*}{GraspNetPC-10 \cite{graspnet, gai}} &Synthetic &Banana, Box, Can, &12,000	&12,000 / -\\
         \cmidrule(lr){2-2}\cmidrule(lr){4-5}
         &Kinect &Camel, Dish, Drill, Mouse, &13,533	&10,973 / 2560\\
         \cmidrule(lr){2-2}\cmidrule(lr){4-5}
         &Realsense &Pear, Scissors, Shampoo &13,258	&10,698 / 2560\\
\bottomrule
    \end{tabular}}
    \label{tab:dataset}
\end{table*}

\begin{figure*}[t]
    \centering
    \begin{subfigure}[t]{0.48\linewidth}
        \centering
        \includegraphics[width=1.05\linewidth]{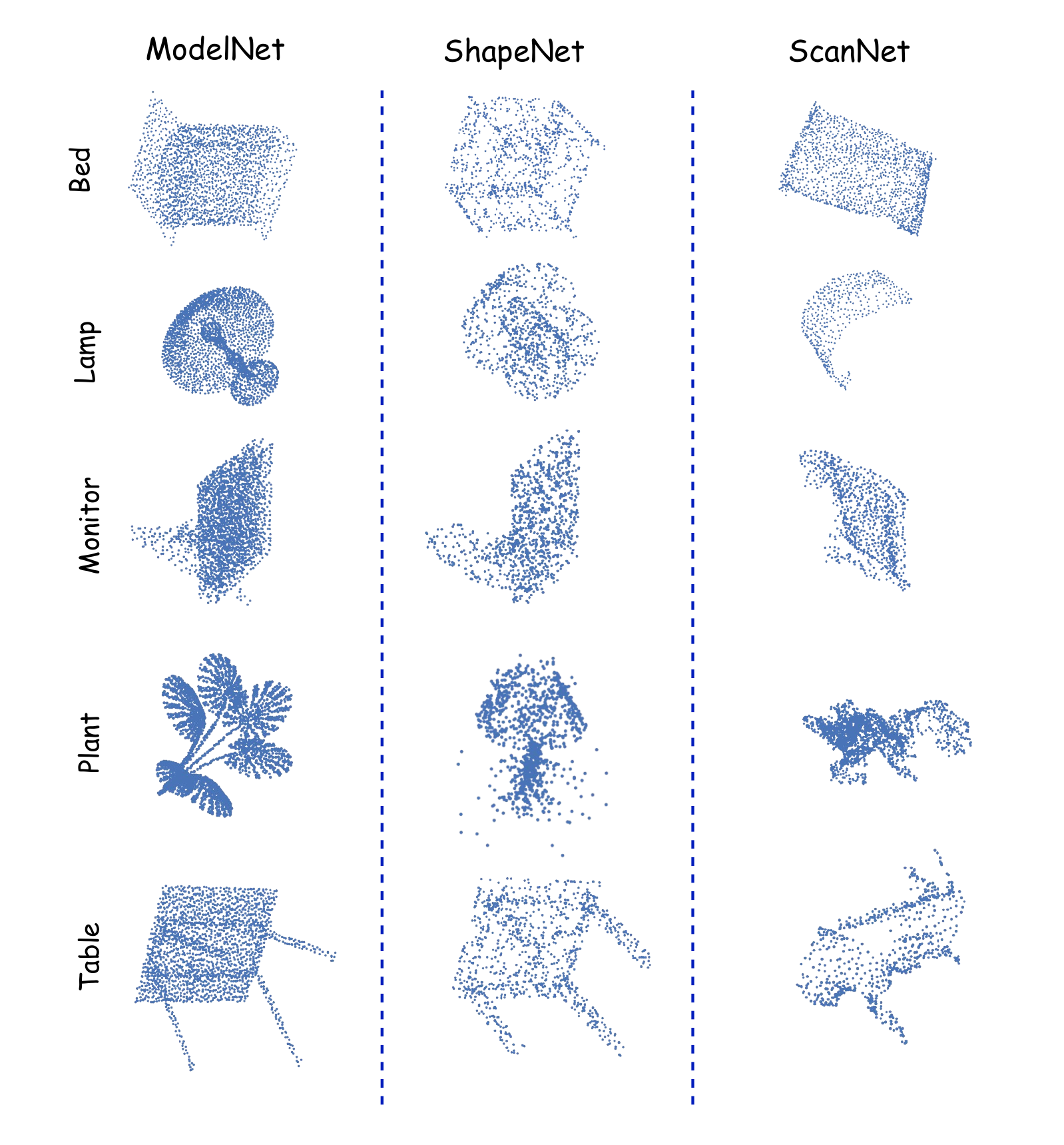}
        \caption{PointDA-10}
        \label{fig:pointda10}
    \end{subfigure}
    \hfill
    \begin{subfigure}[t]{0.48\linewidth}
        \centering
        \includegraphics[width=0.96\linewidth]{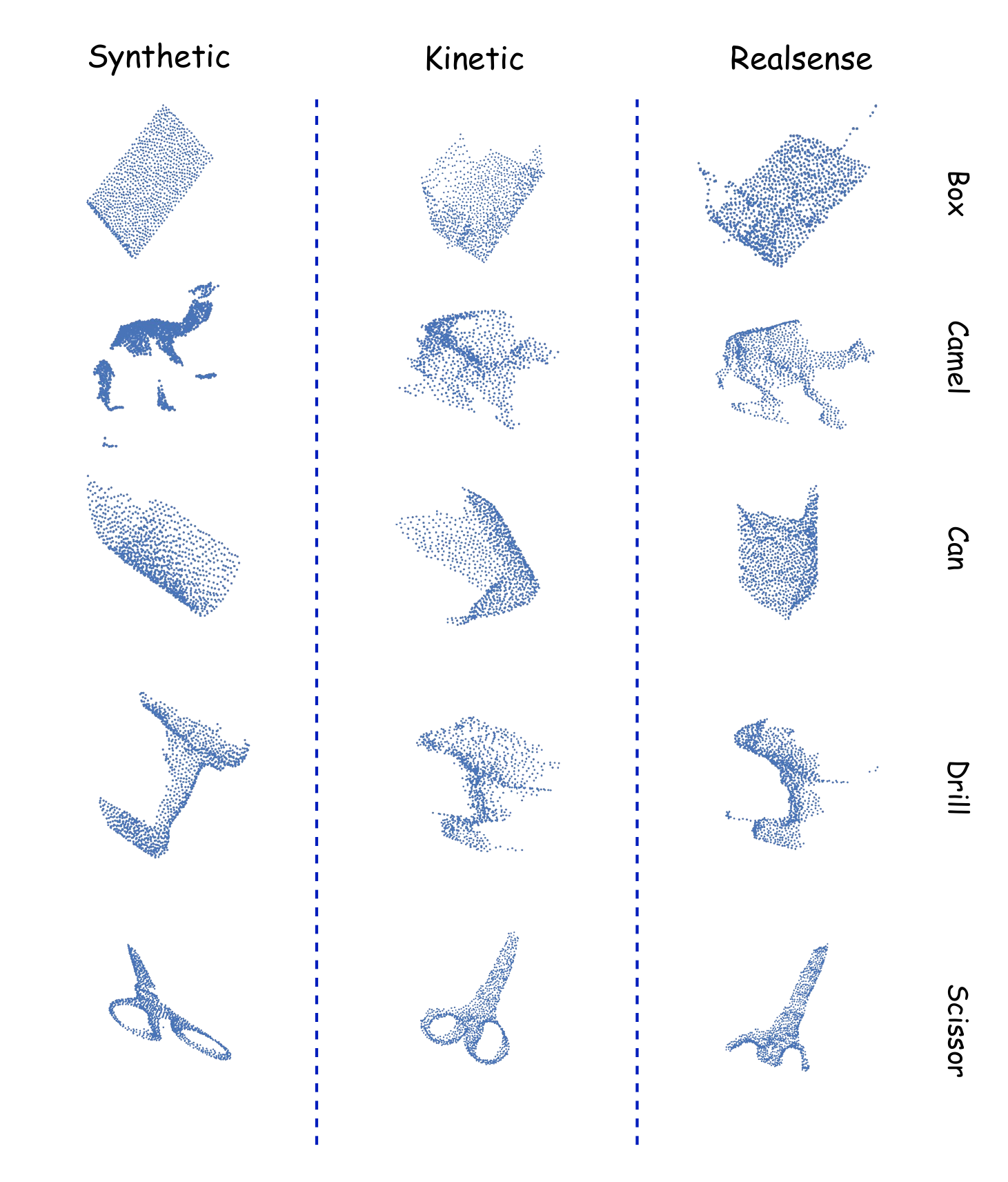}
        \caption{GraspNetPC-10}
        \label{fig:graspnetpc-10}
    \end{subfigure}
    \caption{\textbf{Domain Visualization.} We show the diverse geometry variations across the domains of PointDA-10 and GraspNetPC-10 datasets.}
    \vspace{-0.05cm}
    \label{fig:domainvis}
\end{figure*}

\section{Dataset descriptions}
The PointDA-10 benchmark collects object point clouds from ModelNet40~\cite{modelnet}, ShapeNet~\cite{shapenet}, and ScanNet~\cite{scannet}, covering ten shared object categories. ModelNet-10 (\textbf{M}) dataset contains 4,183 training and 856 testing samples obtained by following online perspective projection \cite{projection} unlike post rendering \cite{postrendering}, i.e., simply projecting each point onto a series of pre-defined image planes to generate scatter depth maps. ShapeNet-10 (\textbf{S}) dataset includes 17,378 training and 2,492 testing point clouds, exhibiting greater structural diversity due to its larger number of object instances and wider geometric variation. ScanNet-10 (\textbf{S$^{*}$}) dataset consists of 6,110 training and 1,769 testing point clouds, containing sensor noise, occlusions, and missing surfaces inherent to reconstructed indoor scenes.

GraspNetPC-10 benchmark is constructed from GraspNet~\cite{graspnet}, a large-scale dataset designed for robotic grasping from raw depth scans and reconstructed CAD models. The point clouds are generated by re-projecting depth maps into 3D space and cropping objects using segmentation masks. Unlike PointDA-10, the point clouds in GraspNetPC-10 are not aligned. This benchmark includes three domains: Synthetic (\textbf{Syn.}), Kinect (\textbf{Kin.}), and Realsense (\textbf{RS.}), corresponding to CAD-rendered depth scans and raw sensor captures from two different depth cameras. The synthetic domain contains 12,000 training samples, while the Kinect and Realsense domains contain 10,973/2,560 and 10,698/2,560 training/testing samples, respectively. Real-world scans from two different depth cameras i.e. Kinect2 and Intel Realsense exhibit domain-specific artifacts, including varying noise patterns, geometric distortions, and missing regions. 

In Figure \ref{fig:domainvis}, we show the diverse geometric variations of different point cloud class objects in synthetic (ModelNet, ShapeNet, Synthetic) and real-world (ScanNet, Kinect, Realsense) environments on both the PointDA-10 and GraspNetPC-10 benchmarks.

\section{LLM attributes generation}
To generate descriptive attributes for each class, we leverage an LLM i.e. GPT-5 \cite{gpt5}. Each class label is passed through a structured instructional prompt, adapted and expanded from the template proposed in~\cite{desc}, as shown in Figure~\ref{fig:attributes}. Specifically, We follow a three-stage pipeline similar to \cite{fedmvp} to construct the attributes integrating two complementary components: a modality-specific prefix template and the attribute set produced by the LLM.
First, we design a prefix that is tailored to the imaging modality of interest i.e. \texttt{``A point cloud object of [class]''}. We then propose a way to enrich this prefix by appending the combination of the LLM-generated attributes on a single sentence using connective phrases such as \texttt{which is a/an''} or \texttt{which has''}, i.e. a single descriptive attribute for each class, which is different from others. This yields a semantically detailed and context-aware prompt that captures both modality information and discriminative visual characteristics. A complete example for the class \texttt{``chair''} is shown in the third row of Figure~\ref{fig:attributes}.

\section{Pseudo-code of CLIPoint3D} 
In Algorithm \ref{pseudocode} we provide the detailed pseudo-codes of training and inference process of CLIPoint3D algorithm.

\begin{figure}[!htbp]
    \centering
    \includegraphics[width=\columnwidth]{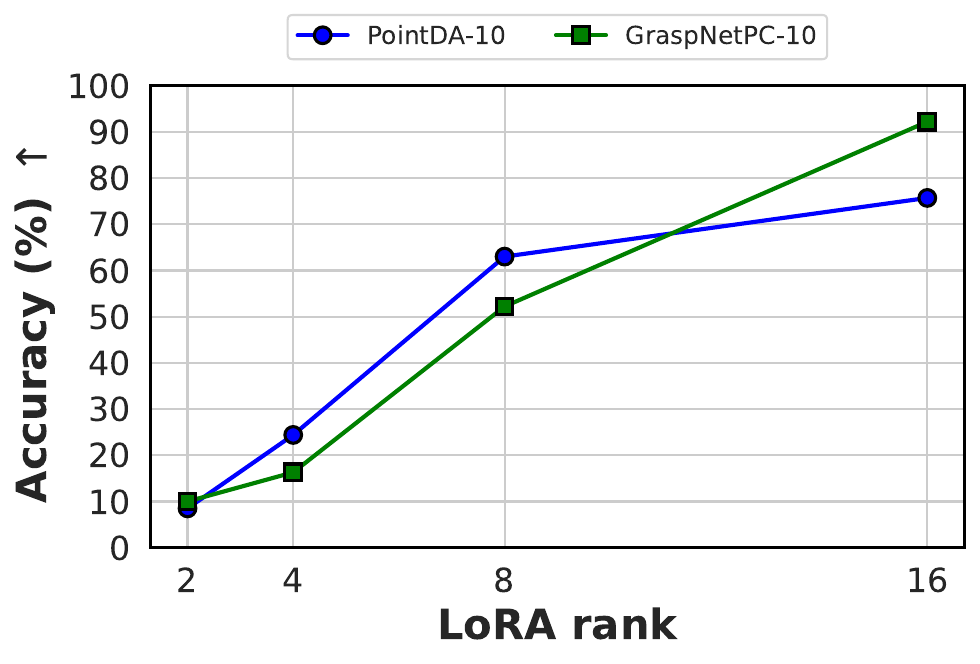}
    \caption{\textbf{Effect of varying LoRA rank.} We report the adaptation performances of CLIPoint3D-V and CLIPoint3D-B on PointDA-10 and GraspNetPC-10 datasets respectively.}
    \label{fig:rank}
\end{figure}

\begin{table*}[t]
    \centering
    \caption{\textbf{Comparison of plug-in UDA methods in CLIP baselines with CLIPoint3D}. We report the adaptation performances on the PointDA-10 benchmark. 
    M: ModelNet, S: ShapeNet, S$^{*}$: ScanNet; $\rightarrow$ indicates the adaptation direction. Best results and second-best results are reported in bold and underlined, respectively.}
    \scalebox{0.8}{
    \begin{tabular}{l|cccccc|c}
    \toprule
        \textbf{Methods} & \textbf{M$\rightarrow$S}  & \textbf{M$\rightarrow$S$^{*}$} & \textbf{S$\rightarrow$M} & \textbf{S$\rightarrow$S$^{*}$} & \textbf{S$^{*}\rightarrow$M} & \textbf{S$^{*}\rightarrow$S} & \textbf{Avg} \\
    \midrule
        ZS-CLIP \cite{clip}  & 46.1& 17.0& 52.0& 17.0& 52.0& 46.1 & 38.4\\ 
        
        CLIP \cite{clip} + DANN \cite{dann} &62.0	&8.6	&77.3	&10.5	&56.7	&50.4	&44.3 \\

        CLIP + CDAN \cite{cdan} &60.9	&7.0	&76.5	&11.0	&56.7	&50.0	&43.7 \\
        CLIP + SCDA \cite{scda} &46.5	&16.2	&51.2	&17.0	&51.8	&46.2	&38.2 \\
        \midrule

        PointCLIP \cite{pointclip} &50.8 &20.9 & 50.1& 20.9& 50.1& 50.8& 40.6\\

        PointCLIP \cite{pointclip} + DANN &55.3	&9.8	&74.2	&14.3	&50.4	&49.7	&42.3  \\
        PointCLIP + CDAN &55.8	&9.2	&72.1	&13.7	&50.9	&49.3	&41.8 \\
        PointCLIP + SCDA &39.2	&17.6	&70.9	&19.8	&67.6	&37.6	&42.1\\
        \midrule

        PointCLIPv2 \cite{pointclipv2} &38.8 & 19.5& 71.6& 19.5& 71.6& 38.8& 43.3\\

        PointCLIP \cite{pointclip} + DANN &46.2	&12.2	&80.4	&13.6	&79.5	&40.6	&45.4  \\
        PointCLIP + CDAN &44.6	&12.8	&75.7	&12.9	&76.8	&41.7	&44.1 \\
        PointCLIP + SCDA &45.9	&12.0	&74.8	&12.5	&77.2	&40.2	&43.8\\
        \midrule

        \cellcolor[gray]{0.93}\textbf{CLIPoint3D-T} &\cellcolor[gray]{0.93}74.4 &\cellcolor[gray]{0.93}9.5 &\cellcolor[gray]{0.93}86.0 &\cellcolor[gray]{0.93}24.1 &\cellcolor[gray]{0.93}50.5 &\cellcolor[gray]{0.93}59.8 &\cellcolor[gray]{0.93}50.7 \\
        
        \cellcolor[gray]{0.93}\textbf{CLIPoint3D-V} &\cellcolor[gray]{0.93}\textbf{84.6} &\cellcolor[gray]{0.93}\textbf{53.5} &\cellcolor[gray]{0.93}\textbf{91.6} &\cellcolor[gray]{0.93}\textbf{55.3} &\cellcolor[gray]{0.93}\textbf{87.9} &\cellcolor[gray]{0.93}\underline{81.3} &\cellcolor[gray]{0.93}\textbf{75.7} \\

        \cellcolor[gray]{0.93}\textbf{CLIPoint3D-B} &\cellcolor[gray]{0.93}\underline{81.5} &\cellcolor[gray]{0.93}\underline{51.9} &\cellcolor[gray]{0.93}\underline{90.3} &\cellcolor[gray]{0.93}\underline{46.6}&\cellcolor[gray]{0.93}\underline{85.2} &\cellcolor[gray]{0.93}\textbf{85.8} &\cellcolor[gray]{0.93}\underline{73.6} \\
    \bottomrule
    \end{tabular}}
    \label{tab:uda}
\end{table*}

\section{Analysis of the LoRA rank}
To understand the influence of the low‐rank decomposition on adaptation quality, we conduct an ablation study on LoRA ranks $2, 4, 8$ \& $16$ using our CLIPoint3D framework. We choose the vision and both encoder variants for the PointDA-10 and GraspNetPC-10 benchmarks respectively. As shown in Figure~\ref{fig:rank}, increasing the LoRA rank consistently improves performance, but the rate of improvement differs markedly between the two datasets.

On PointDA-10, which contains relatively clean synthetic CAD models alongside noisier real scans, accuracy improves steadily from rank 2 to rank 16. The sharp gain from rank 4 to rank 8 indicates that a moderate rank is essential to capture the geometric variability and structural inconsistencies across domains. Beyond rank~8, the improvement becomes more modest, suggesting diminishing returns as representational capacity saturates. Whereas on GraspNetPC-10, whose features are more complex real‐world sensor noise and greater intra-class variation, the benefit of increasing the LoRA rank is even more pronounced. Performance rises from rank~2 to rank~16, with a substantial leap between rank~8 and rank~16. The results show that higher-rank LoRA modules in CLIPoint3D provide the expressive adaptability and generalizability on realistic depth distortions, object incompleteness, and viewpoint variability present in the point cloud domains.

\begin{figure}[!htbp]
    \centering
    \includegraphics[width=\columnwidth]{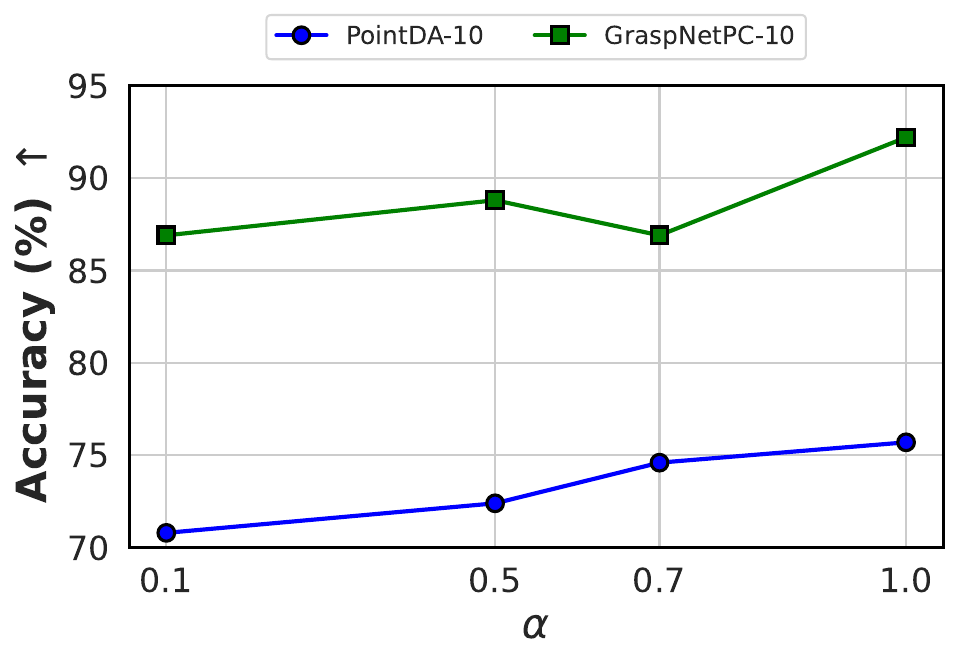}
    \caption{\textbf{Effect of varying $\alpha$ hyperparameter.} We report the adaptation performances of CLIPoint3D-V and CLIPoint3D-B on PointDA-10 and GraspNetPC-10 datasets respectively.}
    \label{fig:alpha}
\end{figure}

\begin{table}[t]
    \centering
    \caption{\textbf{Effect of using various CLIP variants on CLIPoint3D-V}. We report the adaptation performances on the PointDA-10 benchmark.}
    \vspace{-0.2cm}
    \scalebox{0.8}{
    \begin{tabular}{l|ccc}
    \toprule
        \textbf{Methods} & \textbf{ViT-B/16}  & \textbf{ViT-B/32} & \textbf{ViT-L/14} \\
    \midrule

    ModelNet$\rightarrow$ShapeNet & 84.6 &82.4 & 85.7 \\
    ModelNet$\rightarrow$ScanNet & 53.5 &42.7 & 54.4\\
    ShapeNet$\rightarrow$ModelNet & 91.6 &88.3 & 92.1 \\
    ShapeNet$\rightarrow$ScanNet & 55.3 &36.9 & 59.5 \\
    ScanNet$\rightarrow$ModelNet & 87.9 &88.2 & 88.5 \\
    ScanNet$\rightarrow$ShapeNet & 81.3 &73.7 & 82.3 \\
    \midrule
    Average & 75.7 &68.7 & 77.1 \\
    \bottomrule
    \end{tabular}}
    \label{tab:backbone}
\end{table}

\section{Conventional plug-in UDA methods in CLIP baselines} 
In Table \ref{tab:uda}, we evaluate the effect of integrating standard UDA techniques e.g. DANN \cite{dann}, CDAN \cite{cdan}, and SCDA \cite{scda} into our CLIP-based baselines (Zs-CLIP, PointCLIP \& PointCLIPv2) to train them for point-cloud UDA task. While these conventional UDA methods can bring modest improvements in certain cross-domain transfers, their gains are inconsistent and often fail to fully bridge the domain gap inherent in 3D point cloud data. To be noted, we have just added a learnable adapter on top of the frozen visual features of the vision encoder similar to \cite{clipadapter}, while keeping both the encoders entirely frozen. While the CLIP-based methods improve adaptation after pluggin on the UDA methods, but still underperform compared to CLIPoint3D variants. It highlights the limitations of applying traditional 2D-centric UDA strategies directly to CLIP-based 3D recognition. The results of Table \ref{tab:uda} suggest that while conventional UDA provides some benefits, more specialized adaptation strategies are necessary to consistently leverage the cross-modal representations of CLIP and achieve robust performance across diverse 3D domains.

\section{Effect of $\alpha$ hyperparameter} 
We investigate the influence of the $\alpha$ hyperparameter in the total loss function (\textit{Eq. 13 of main paper}) in our proposed CLIPoint3D. As shown in Fig. \ref{fig:alpha}, varying $\alpha$ affects the trade-off between different components of the loss and consequently the adaptation performance. For PointDA-10, increasing $\alpha$ leads to a steady improvement, indicating that a higher weight on the corresponding loss term better guides the model for cross-domain alignment. In contrast, GraspNetPC-10 exhibits a more varied trend, with performance peaking at intermediate and higher $\alpha$ values, suggesting that overly small or excessively large weighting can underutilize certain loss components.

\begin{table*}[t]
    \centering
    \caption{\textbf{Effect of using different LLMs on CLIPoint3D-B for attributes generation}. We report the adaptation performances on the GraspNetPC-10 benchmark. and choose a snippet of attributes of class `drill machine'.}
    \scalebox{0.67}{
    \begin{tabular}{l|c|cccc|c}
    \toprule
        \textbf{Methods} & \textbf{Attributes Snippet} & \textbf{Syn.$\rightarrow$Kin.}  & \textbf{Syn$\rightarrow$RS.} & \textbf{Kin.$\rightarrow$RS.} & \textbf{RS.$\rightarrow$Kin.} &\textbf{Avg} \\
    \midrule

Handcrafted & ``A point cloud object of'' &93.5 &81.6 &77.9 &94.8 &87.0 \\
Llama-3.2-3B \cite{llama3}  & ``Compact, rectangular structure with a flat surface'' &95.5 &84.1 &84.3 &95.6 &89.9 \\ 
Qwen2.5-14B \cite{qwen2}  &``Cylindrical main body with handle and trigger'' &95.0 &87.5 &83.2 &95.4 &90.3 \\
Phi-4 \cite{phi4} & ``Irregular mechanical shape'' &95.8 &83.2 &79.6 &92.3 &87.7 \\
GPT-5 \cite{gpt5} & ``Elongated body with a cylindrical or pistol-like grip'' &\textbf{96.5} &\textbf{89.3} &\textbf{86.8} &\textbf{96.2} &\textbf{92.2} \\

    \bottomrule
    \end{tabular}}
    \label{tab:llm_effect}
\end{table*}

\section{Influence of the prompt length}
\begin{figure}[!htbp]
    \centering
    \includegraphics[width=\columnwidth]{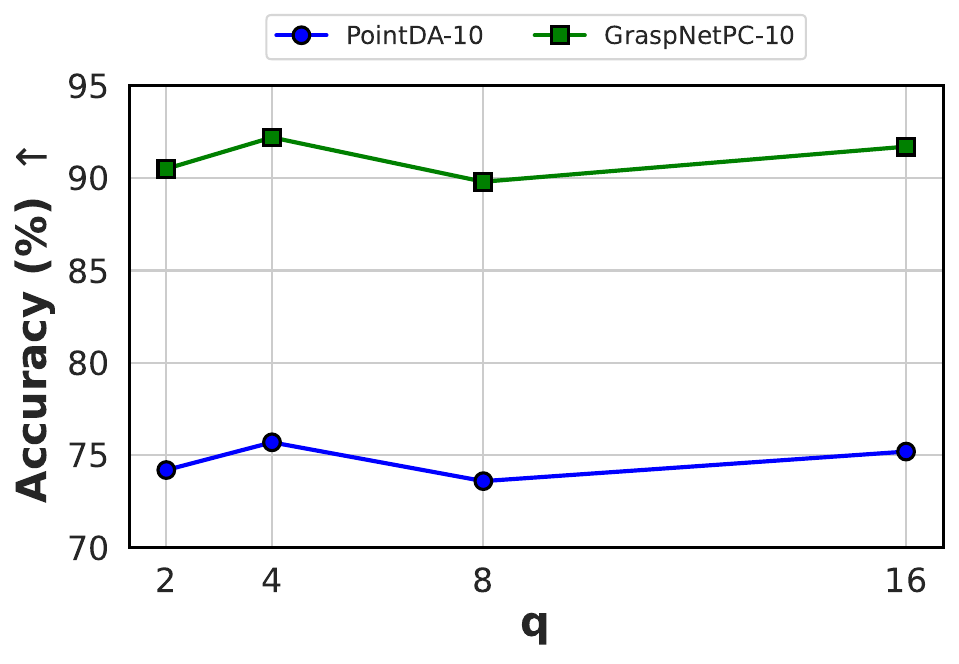}
    \caption{\textbf{Effect of varying length of $\mathbf{q}$.} We report the adaptation performances of CLIPoint3D-V and CLIPoint3D-B on PointDA-10 and GraspNetPC-10 datasets respectively.}
    \label{fig:length}
\end{figure}

We analyze the effect of the length of shared prompt $\mathbf{q}$ on CLIPoint3D. As shown in Fig. \ref{fig:length}, varying $\mathbf{q}$ affects the UDA performances. We choose the vision and both encoder variants for the PointDA-10 and GraspNetPC-10 benchmarks respectively. For PointDA-10, the performance remains relatively stable across different $\mathbf{q}$ values, indicating that CLIPoint3D is robust to moderate changes in prompt length. In contrast, GraspNetPC-10 exhibits slight fluctuations, with intermediate values of $\mathbf{q}$ yielding the best results.

\section{Impact of CLIP variants}
We investigate how different CLIP backbone architectures affect the performance of our CLIPoint3D method and choose the vision variant to study on PointDA-10 benchmark. Table \ref{tab:backbone} compares results using ViT-B/16, ViT-B/32, and ViT-L/14. Overall, larger frozen ViT backbones tend to provide stronger feature representations, leading to improved domain adaptation performance. Specifically, ViT-L/14 achieves the highest average accuracy, benefiting from its larger model capacity and fine-grained patch representation. ViT-B/16 offers a strong trade-off between efficiency and performance, outperforming ViT-B/32 in most cases despite similar model sizes. It indicates that the choice of backbone has a significant impact on the effectiveness of cross-domain alignment, while showcasing CLIP has the extreme potential of capturing better the nuances of 3D point cloud distributions.

\begin{algorithm*}[t!]
\caption{CLIPoint3D algorithm}
\label{pseudocode}
\begin{algorithmic}[1]
\Require Training data: source domain $\mathcal{D}^{\mathcal{S}_l} = \{(P_i^{\mathcal{S}_l}, y_i^{\mathcal{S}_l})\}_{i=1}^{N_{\mathcal{S}_l}} = \{(x_{i,m}^{\mathcal{S}_l}, y_i^{\mathcal{S}_l})\}$ and target domain $\mathcal{D}^{\mathcal{T}_u} = \{P_j^{\mathcal{T}_u}\}_{j=1}^{N_{\mathcal{T}_u}} = \{x_{j,m}^{\mathcal{T}_u}\}$, $\mathcal{E}_t$, $\mathcal{E}_v$ \& $\mathcal{E}_{3D}$.

\Procedure{Training Objective}{}
\State Generate the attributes of set of classes, $\mathcal{C} = \{c_{k}\}^{K}_{k=1}$, by a $\texttt{LLM}$, and extract $\mathbf{T}^{\text{llm}}$ using Eq.2.
\State Generate $M$ 2D projected depth maps for each 3D point cloud sample.
\State Initialize a random prompt vector of length $l$ from a Gaussian distribution.
\If{$n = 1$} \Comment{Given total $\mathcal{N}$ epochs}
\For{\( i \) $\gets$ 0 \textbf{to} \( K \)} \Comment{Given $K$ iterations}
     \State Generate textual prompts $\mathbf{P}_{t}(\mathbf{T}^{\text{llm}}, \mathbf{p})$ using Eq.3, and extract textual embeddings $\mathbf{T}$ from $\mathcal{E}_t$.
     \State Generate source and target visual prompts ($\mathbf{P}_{v}^{\mathcal{S}}$ \& $\mathbf{P}_{v}^{\mathcal{T}}$) separately using Eq.4.
     \State Extract visual embeddings $\mathbf{I}_{\mathcal{S}_l}$ and $\mathbf{I}_{\mathcal{T}_u}$ from $\mathcal{E}_v^{\mathcal{S}}$ \& $\mathcal{E}_v^{\mathcal{T}}$ respectively.
     \State Do PEFT adaptation of text / vision / both encoder(s).
     \State Select the views of minimum entropy using Eq.5 and calculate the final prediction probability of a point cloud sample using Eq.6.
     \State Calculate $\mathbf{L}_{\mathrm{ce}}$, $\mathbf{L}_{\mathrm{ortho}}$, $\mathbf{L}_{\mathrm{OT}}$ \& $\mathbf{L}_{\mathrm{conf}}$ using Eq.13.
     \State Append the source batch probabilities $p_{\mathcal{S}_l}$ in a list.
\EndFor
\State Save all $p_{\mathcal{S}_l}$ and calculate uncertainty-weighted source class prototypes using Eq.7.    
\EndIf
\For{\( n \) $\gets$ 2 \textbf{to} \( \mathcal{N} \)}
\State Repeat the procedure of steps 7-15.
\State Calculate $\mathbf{L}_{\mathrm{proto}}$ using the source prototypes from the $(n-1)$-th epoch and Eq.10.
\State Calculate $\mathbf{L}_{\mathrm{total}}$ using Eq.13.
\EndFor

\EndProcedure
\Procedure{Inference}{}
\State Consider all test samples of target domain $\mathcal{D}^{\mathcal{T}_u}$ in the source dataloader and calculate $\texttt{top-1}$ accuracy by selecting the class of maximum $p_{\mathcal{T}_u}$. 
\EndProcedure
\end{algorithmic}
\end{algorithm*}

\section{Effect of various LLMs}
We examine the impact of using different LLMs to generate semantic attributes for our method CLIPoint3D and choose the both encoder variant to study on GraspNetPC-10 benchmark.. Table \ref{tab:llm_effect} summarizes the adaptation performance when attributes are derived from GPT-5 \cite{gpt5}, Llama-3.2-3B \cite{llama3}, Qwen2.5-14B \cite{qwen2}, and Phi-4 \cite{phi4}. Across all adaptation scenarios, GPT-5 consistently produces the most informative attributes, leading to the highest average performance. While other LLMs such as Llama-3.2-3B and Qwen2.5-14B yield competitive results in certain domain pairs, their overall effectiveness is slightly lower and more variable. The results highlight that the quality and expressiveness of the generated attributes significantly influence cross-domain alignment, emphasizing the importance of selecting a capable LLM for robust 3D domain adaptation.

\end{document}